\PassOptionsToPackage{unicode}{hyperref}
\PassOptionsToPackage{hyphens}{url}
\PassOptionsToPackage{dvipsnames,svgnames,x11names}{xcolor}
\documentclass[
]{article}

\usepackage{amsmath,amssymb}
\usepackage{iftex}
\ifPDFTeX
  \usepackage[T1]{fontenc}
  \usepackage[utf8]{inputenc}
  \usepackage{textcomp} 
\else 
  \usepackage{unicode-math}
  \defaultfontfeatures{Scale=MatchLowercase}
  \defaultfontfeatures[\rmfamily]{Ligatures=TeX,Scale=1}
\fi
\usepackage{lmodern}
\IfFileExists{upquote.sty}{\usepackage{upquote}}{}
\IfFileExists{microtype.sty}{
  \usepackage[]{microtype}
  \UseMicrotypeSet[protrusion]{basicmath} 
}{}
\makeatletter
\@ifundefined{KOMAClassName}{
  \IfFileExists{parskip.sty}{%
    \usepackage{parskip}
  }{
    \setlength{\parindent}{0pt}
    \setlength{\parskip}{6pt plus 2pt minus 1pt}}
}{
  \KOMAoptions{parskip=half}}
\makeatother
\usepackage{xcolor}
\makeatletter
\ifx\paragraph\undefined\else
  \let\oldparagraph\paragraph
  \renewcommand{\paragraph}{
    \@ifstar
      \xxxParagraphStar
      \xxxParagraphNoStar
  }
  \newcommand{\xxxParagraphStar}[1]{\oldparagraph*{#1}\mbox{}}
  \newcommand{\xxxParagraphNoStar}[1]{\oldparagraph{#1}\mbox{}}
\fi
\ifx\subparagraph\undefined\else
  \let\oldsubparagraph\subparagraph
  \renewcommand{\subparagraph}{
    \@ifstar
      \xxxSubParagraphStar
      \xxxSubParagraphNoStar
  }
  \newcommand{\xxxSubParagraphStar}[1]{\oldsubparagraph*{#1}\mbox{}}
  \newcommand{\xxxSubParagraphNoStar}[1]{\oldsubparagraph{#1}\mbox{}}
\fi
\makeatother

\usepackage{longtable,booktabs,array}
\usepackage{calc} 
\usepackage{etoolbox}
\makeatletter
\patchcmd\longtable{\par}{\if@noskipsec\mbox{}\fi\par}{}{}
\makeatother
\IfFileExists{footnotehyper.sty}{\usepackage{footnotehyper}}{\usepackage{footnote}}
\makesavenoteenv{longtable}
\usepackage{graphicx}
\makeatletter
\newsavebox\pandoc@box
\newcommand*\pandocbounded[1]{
  \sbox\pandoc@box{#1}%
  \Gscale@div\@tempa{\textheight}{\dimexpr\ht\pandoc@box+\dp\pandoc@box\relax}%
  \Gscale@div\@tempb{\linewidth}{\wd\pandoc@box}%
  \ifdim\@tempb\p@<\@tempa\p@\let\@tempa\@tempb\fi
  \ifdim\@tempa\p@<\p@\scalebox{\@tempa}{\usebox\pandoc@box}%
  \else\usebox{\pandoc@box}%
  \fi%
}
\def\fps@figure{htbp}
\makeatother

\usepackage{booktabs}
\usepackage{multirow}
\usepackage{longtable}
\usepackage[section]{placeins}
\usepackage{float}
\usepackage{array}
\usepackage{ragged2e}
\usepackage{enumitem}
\usepackage{graphicx}
\newcolumntype{L}[1]{>{\RaggedRight\arraybackslash}p{#1}}
\usepackage{arxiv}
\usepackage{orcidlink}
\usepackage{amsmath}
\usepackage[T1]{fontenc}
\makeatletter
\@ifpackageloaded{caption}{}{\usepackage{caption}}
\AtBeginDocument{%
\ifdefined\contentsname
  \renewcommand*\contentsname{Table of contents}
\else
  \newcommand\contentsname{Table of contents}
\fi
\ifdefined\listfigurename
  \renewcommand*\listfigurename{List of Figures}
\else
  \newcommand\listfigurename{List of Figures}
\fi
\ifdefined\listtablename
  \renewcommand*\listtablename{List of Tables}
\else
  \newcommand\listtablename{List of Tables}
\fi
\ifdefined\figurename
  \renewcommand*\figurename{Figure}
\else
  \newcommand\figurename{Figure}
\fi
\ifdefined\tablename
  \renewcommand*\tablename{Table}
\else
  \newcommand\tablename{Table}
\fi
}
\@ifpackageloaded{float}{}{\usepackage{float}}
\floatstyle{ruled}
\@ifundefined{c@chapter}{\newfloat{codelisting}{h}{lop}}{\newfloat{codelisting}{h}{lop}[chapter]}
\floatname{codelisting}{Listing}

\makeatother
\makeatletter
\@ifpackageloaded{caption}{}{\usepackage{caption}}
\@ifpackageloaded{subcaption}{}{\usepackage{subcaption}}
\makeatother

\usepackage{bookmark}

\IfFileExists{xurl.sty}{\usepackage{xurl}}{} 
\hypersetup{
  pdftitle={PyPottery: an AI-powered end-to-end suite for pottery processing and publication},
  pdfauthor={Lorenzo Cardarelli},
  pdfkeywords={AI, ML, Material Culture, Pottery, Computer
Vision, LLMs},
  colorlinks=true,
  linkcolor={blue},
  filecolor={Maroon},
  citecolor={Blue},
  urlcolor={Blue},
  pdfcreator={LaTeX via pandoc}}

\newcommand{\runninghead}{A Preprint }
\renewcommand{\runninghead}{PyPottery }
\title{PyPottery: an AI-powered end-to-end suite for pottery processing
and publication\thanks{Work on this manuscript was conducted while the
author was affiliated with Sapienza Università di Roma, Department of
Antiquities, and Parco Archeologico del Colosseo, Via in Miranda 5,
00186 Rome, Italy.}}
\author{\textbf{Lorenzo Cardarelli}\\Seminars für Ur- und
Frühgeschichte\\Georg-August-Universität
Göttingen\\Göttingen,\ 37073\\\\ISMed-CNR -- Istituto di Studi sul
Mediterraneo\\Naples,\ 80134\\\href{mailto:lorenzo.cardarelli@uni-goettingen.de}{lorenzo.cardarelli@uni-goettingen.de}}
\date{}
\begin{document}
\maketitle
\begin{abstract}
The study of ceramic materials constitutes a cornerstone of
archaeological research, yet the post-production workflow for pottery
documentation remains labor-intensive and creates significant
publication bottlenecks. This paper presents PyPottery, an open-source,
AI-powered suite designed to semi-automate the complete ceramic
documentation pipeline. The suite comprises four integrated modules:
PyPotteryScan for automated image extraction and handwriting
recognition; PyPotteryInk for automatic inking of pencil drawings;
PyPotteryTrace for semantically-aware vectorization; and PyPotteryLayout
for automated layout generation. Evaluated on 50 hand-drawn sheets
containing 240 pottery drawings from the Terramara di Montale (Italy),
the framework achieved substantial time savings confirmed by usability
study participants, who reported a median perceived speedup of
40\(\times\) over traditional workflows (range:
17.5\(\times\)--120\(\times\)). These results highlight the potential of
AI-assisted tools in archaeological documentation, while the paper
addresses the strategic redistribution of cognitive labor toward
augmentation rather than automation.
\end{abstract}
{\bfseries \emph Keywords}
\def\sep{\textbullet\ }
AI \sep ML \sep Material Culture \sep Pottery \sep Computer Vision \sep 
LLMs

\section{Introduction}\label{introduction}

\subsection{Role of ceramics and post-production
bottleneck}\label{role-of-ceramics-and-post-production-bottleneck}

The study of material culture plays a fundamental role in understanding
ancient civilizations and cultures. Among the myriad artifacts recovered
from archaeological excavations, pottery holds a position of
unparalleled importance. Its ubiquity across time and space, combined
with its durability and stylistic variability, makes it one of the most
informative categories of archaeological evidence. Ceramic assemblages
are crucial for establishing chronological sequences, understanding
technological traditions, and reconstructing patterns of production,
distribution, and consumption \cite{ref1,ref2,ref3}. The detailed
analysis of pottery shapes, fabrics, and decorations allows
archaeologists to address a wide range of research questions, from the
daily lives of individuals to the large-scale dynamics of ancient
economies and societies. With the advent of artificial intelligence (AI)
in archaeology \cite{ref4,ref5,ref6}, new opportunities have emerged to
develop tools that facilitate the analysis \cite{ref7,ref8} and
cataloging of existing data \cite{ref9,ref10}. In the broader field of
digital archaeology, IT-related approaches to pottery documentation have
a long history (early works: \cite{ref11,ref12,ref13,ref14}, recent
reviews: \cite{ref15}), with the \emph{Laser Aided Profiler} (LAP)
\cite{ref16} playing a pioneering role and gaining popularity in recent
years. Despite these technological advances, manual drawing remains the
prevailing documentary standard, especially in university contexts where
it constitutes a fundamental part of archaeologists' training
\cite{ref17}, valued not merely as a representational technique but as
an interpretive act.

However, this reliance on manual methods introduces a significant
bottleneck in the archaeological workflow, particularly in the
subsequent stages of digital processing and publication. The standard
post-production workflow involves labor-intensive steps: scanning
hand-drawn sheets, manually cropping individual drawings, transcribing
associated textual data (context, inventory number, stratigraphic unit)
into a database, digitally inking drawings using vector graphics
software (e.g., Adobe Illustrator, Inkscape), and laying out materials
for publication. This entire sequence requires considerable time
investment and specialized software expertise. The post-production
bottleneck may lead to the creation of a ``data graveyard'' where
valuable archaeological information remains unpublished and inaccessible
to the wider scholarly community.

The advent of machine learning (ML) and large language models (LLMs)
presents a transformative opportunity for archaeology
\cite{ref4,ref10,ref18}. Emerging AI-assisted tools address various
aspects of archaeological documentation \cite{ref9,ref10,ref19}, and
tools like PyPotteryInk \cite{ref20} have demonstrated successful
automation of specific tasks (ceramic drawing inking via deep learning).
However, the development of increasingly powerful LLMs, coupled with a
more sophisticated understanding of their application, enables the
creation of integrated, end-to-end solutions. The repetitive and
mechanical nature of post-production tasks makes them ideally suited for
AI-driven optimization. However, existing tools address only isolated
phases of the workflow, leaving a critical gap: the absence of an
integrated solution capable of managing the entire post-production
pipeline within a unified, accessible framework. In fields like
archaeological documentation, where adoption barriers are high and users
are domain experts rather than software developers, a well-integrated,
empirically validated suite constitutes a scientific contribution in its
own right, distinct from the individual algorithmic advances it
incorporates.

\subsection{Related works}\label{related-works}

Several tools have recently addressed specific aspects of archaeological
pottery documentation using AI. PyPotteryInk \cite{ref20} and
PyPotteryLens \cite{ref9} demonstrated the automation of specific
documentation tasks --- inking and digitization respectively --- but
addressed only isolated phases of the workflow. Klein et
al.~\cite{ref10} proposed an AI-assisted workflow for object detection
and data collection from archaeological catalogues, while Pang et
al.~\cite{ref21} developed a vision-language model (VLM)-based pipeline
for automated collection of multimodal data from published ceramic
catalogs, combining document localization, block comprehension, and
image-text matching. Tang et al.~\cite{ref22} introduced
PDFChatAnnotator, a human-LLM collaborative tool for annotating
multimodal data from PDF catalogs, sharing PyPotteryScan's emphasis on
structured metadata extraction with human oversight. In the domain of
line drawing generation, Xue et al.~\cite{ref23} proposed a generative
framework for producing archaeological line drawings from artifact
photographs using DiT architecture and LoRA fine-tuning, while Wang et
al.~\cite{ref24} developed Lineart, a knowledge-guided diffusion model
for appearance transfer in design drawings --- both approaches
complementary to but distinct from PyPotteryInk's pencil-to-ink
translation paradigm. Alongside these research prototypes,
community-driven platforms such as CeraLoop
(\url{https://ceraloop.lad-sapienza.it}) are emerging as collaborative
infrastructures for ceramic documentation, indicating a growing
disciplinary demand for shared, AI-assisted resources. Vectorization, by
contrast, has largely been left to general-purpose tools such as Potrace
(\url{https://potrace.sourceforge.net/}) and VTracer
(\url{https://github.com/visioncortex/vtracer}), which were designed for
generic raster-to-vector conversion and are not sensitive to the
semantic structure of archaeological drawings (see
Section~\ref{sec-trace}). Despite this growing landscape, no existing
solution integrates scanning, OCR, parsing, inking, vectorization, and
layout generation into a unified, archaeologist-oriented framework ---
the gap PyPottery addresses.

\section{Research Aims}\label{research-aims}

This paper presents PyPottery, an open-source, AI-powered suite designed
to semi-automate the ceramic documentation pipeline. The objectives are:
(1) detail the tool's architecture, including its four integrated
modules (PyPotteryScan, PyPotteryInk, PyPotteryTrace, PyPotteryLayout)
and their underlying AI technologies; (2) assess the framework's
performance on a real-world dataset of 240 pottery drawings, measuring
processing time, recognition accuracy, and output quality; (3) evaluate
the tool's practical applicability through a formal usability study with
domain experts; (4) discuss the application of an AI-assisted software
in the context of archaeological documentation.

\section{Materials and Methods}\label{sec-methods}

The complete PyPottery workflow is illustrated in
Fig.~\ref{fig-workflow}, which serves as a structural reference for this
entire section. Each module described below corresponds to a specific
block in the figure, and the reader is invited to refer back to it
throughout.

\subsection{The Traditional Post-Production
Workflow}\label{the-traditional-post-production-workflow}

The traditional methodology for processing hand-drawn archaeological
illustrations follows a well-established, albeit laborious, three-phase
workflow. A representative example of the hand-drawn sheets addressed by
this workflow, containing both artifact drawings and handwritten
metadata, is provided in Supplementary materials, Fig. S1.

\begin{enumerate}
\def\labelenumi{\arabic{enumi}.}
\item
  \textbf{Scanning and cataloging of hand-drawn sheets}: The process
  begins with the digitization of the original paper sheets, which often
  contain multiple artifact drawings and their associated textual
  metadata (e.g., context name, inventory number, stratigraphic unit).
  These sheets are scanned and each artifact drawing is cropped into a
  separate image file, typically in a common raster format like JPG or
  PNG. Simultaneously, the handwritten or typed text accompanying each
  drawing is parsed and manually entered into a structured database.
\item
  \textbf{Inking/vectorization of drawings}: Once the individual
  drawings are isolated, they undergo a process of digital inking. This
  involves using vector graphics software to trace over the pencil lines
  of the original drawing, creating a clean, crisp, and scalable vector
  representation. This is one of the most time-intensive steps in the
  entire workflow.
\item
  \textbf{Layout and final publication}: The final phase involves
  assembling the inked drawings and their associated metadata into a
  publication-ready layout. This is typically done using desktop
  publishing software, where the archaeologist must manually arrange the
  illustrations on pages according to the methodological requirements of
  the publication (e.g., comparative tables, catalogues, etc.).
\end{enumerate}

\subsection{The Motivation for a Human-in-the-Loop
Framework}\label{the-motivation-for-a-human-in-the-loop-framework}

PyPottery is conceived as a modular, human-in-the-loop framework that
retraces and streamlines the traditional post-production phases through
open-source and open-weights AI models, in order to accelerate,
standardize, and democratize the entire process. Within the proposed
framework, point 1 is handled by PyPotteryScan, point 2 is managed by
PyPotteryInk and PyPotteryTrace, while point 3 is supported by
PyPotteryLayout (Table~\ref{tbl-software}).

This mapping, however, is not merely a technical correspondence: the
decision to intervene at each of these points (and the form that
intervention takes) rests on a set of underlying premises that it is
worth making explicit. The structural motivations underlying this
approach derive from a combination of practical, methodological, and
scientific needs. (1) The manual drawing phase remains \emph{currently}
irreplaceable, representing the state-of-the-art in ceramic
archaeological documentation. (2) The archaeological community possesses
an enormous quantity of legacy material (sheets and drawings created
over decades of research activity) which constitutes a valuable
information heritage but is often not digitized or not structured.
Abandoning such material would be uneconomical and scientifically
damaging; instead, tools capable of valorizing it and integrating it
into contemporary digital workflows with minimal effort are needed. (3)
Most currently used software solutions are commercial products that
limit accessibility and scientific reproducibility. An open-source tool
like PyPottery addresses this critical issue by offering transparency of
code and processes, execution on consumer hardware, interoperability
with common and standard file formats, adaptability to the diverse needs
of laboratories and research projects, and active participation of the
scientific community in the tool's development and continuous
improvement. (4) AI open models can be modified, and retrained based on
new datasets, thus enabling collective and shared knowledge growth.

\begin{figure}

\centering{

\includegraphics[width=0.55\linewidth,height=\textheight,keepaspectratio]{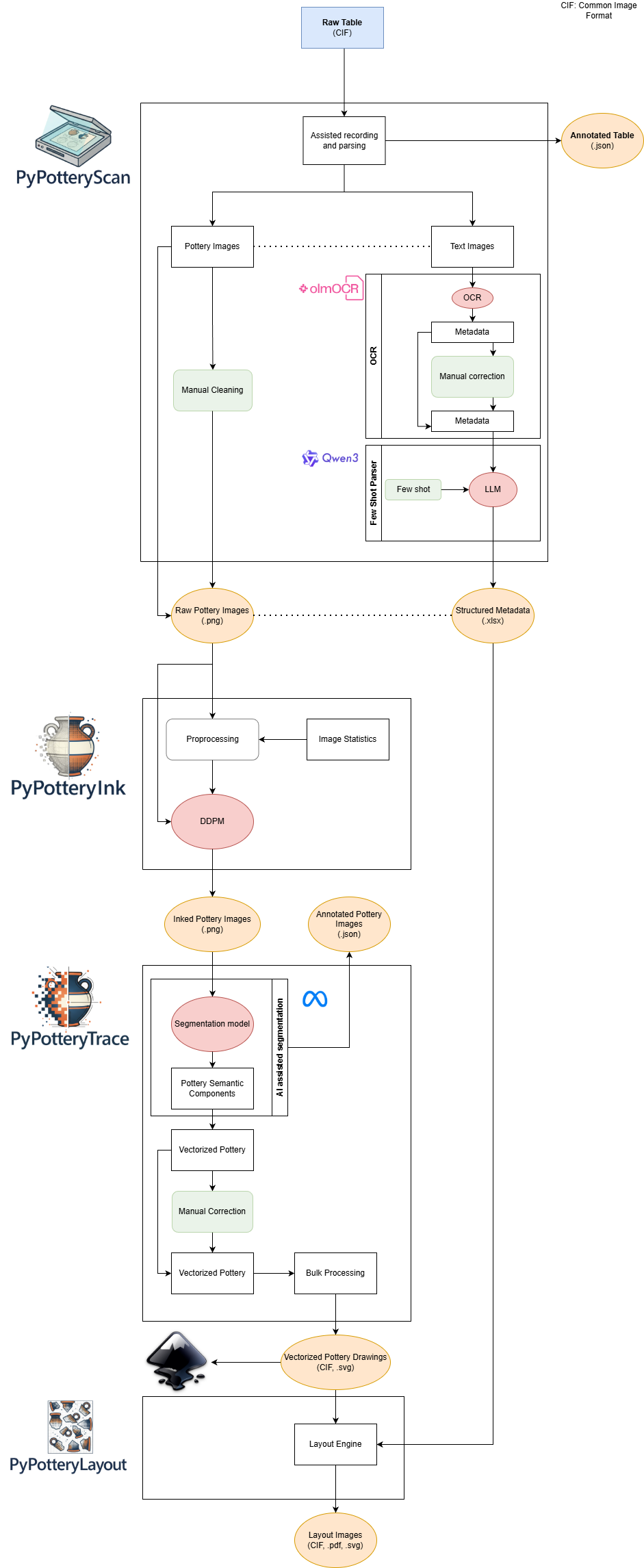}

}

\caption{\label{fig-workflow}The diagram shows the complete PyPottery
workflow. Each block identifies a specific module and its function
within the overall process. Orange boxes indicate outputs and green
boxes indicate manual processing checks.}

\end{figure}%

\begin{table}

\caption{\label{tbl-software}The main PyPottery modules, functions and IT technologies.}

\centering{

[H]
\centering

\small
\begin{tabular}{L{2.3cm}L{3.3cm}L{2.6cm}L{3.8cm}}
\toprule
\textbf{Module} & \textbf{Function} & \textbf{Corresponding Traditional Phase} & \textbf{Key AI Technologies} \\
\midrule
PyPotteryScan & Automated scanning, image cropping, and metadata extraction from sheets. & 1. Scanning and Cataloging & LLMs (olmOCR, Qwen3) for handwriting recognition and structured parsing. \\
\addlinespace
PyPotteryInk & Automatic inking of raster drawings. & 2. Inking/Vectorization & Deep learning models for image-to-image translation. \\
\addlinespace
PyPotteryTrace & Semantically-aware vectorization of inked drawings. & 2. Inking/Vectorization & Segmentation models (SAM~2) and custom vectorization algorithms. \\
\addlinespace
PyPotteryLayout & Automated generation of publication layouts. & 3. Layout and Publication & Spatial optimization algorithms. \\
\bottomrule
\end{tabular}

}

\end{table}%

\subsection{Case Study: Pottery from Terramara di
Montale}\label{case-study-pottery-from-terramara-di-montale}

To test and validate the entire process, ceramic material from the
excavations of the Terramara di Montale (Modena, Italy) \cite{ref25} was
selected as a case study. The traditional documentation workflow for
this material involved the creation of hand-drawn sheets, which included
both artifact drawings and associated textual metadata such as inventory
numbers, stratigraphic units, and excavation year. The drawings exhibit
high graphical quality and follow consistent representational
conventions. However, the accompanying textual data is handwritten in
cursive Italian, presenting challenges for automated recognition.
Typical annotations include stratigraphic indicators (e.g., `US',
`Taglio' -- Italian words for ``Stratigraphic Unit'' and ``Cut'') and
alphanumeric inventory codes. All hand-drawn sheets were digitized at a
resolution of 300 DPI to ensure optimal input quality.

The material from Montale is substantial, and for this work, a random
sample of 50 hand-drawn sheets was selected (Supplementary materials,
Fig. S1). As the archaeological material is currently unpublished, the
full dataset cannot be made publicly available at this stage.

\subsection{Automatic Scanning and Cataloging with
PyPotteryScan}\label{automatic-scanning-and-cataloging-with-pypotteryscan}

PyPotteryScan automates the extraction and pairing of images and textual
information from scanned sheets. The system is structured into two main
phases: assisted recording/parsing and automated content extraction.

\subsubsection{Assisted Recording and Sheet
Annotation}\label{assisted-recording-and-sheet-annotation}

The first phase involves an assisted recording process, in which the
user manually identifies, through a dedicated GUI, the boxes containing
the drawings and their associated text (Fig. \ref{fig-scan_gui}a). This
semi-automatic approach is necessitated by the significant variability
in the arrangement and format of textual information present in
hand-drawn sheets. However, the system is designed with a progressive
improvement perspective: during the recording process, the coordinates
of text boxes and drawings are saved in JSON format, creating an
annotated dataset that can be used in the future for training specific
object detection models for this type of archaeological documentation.
Once the boxes containing the drawings have been identified, the images
can be extracted and, if necessary, cleaned through editing tools
integrated into the GUI (e.g., removal of annotations or disturbing
elements using an ``eraser'' tool). This step produces individual image
files in common image format for each artifact, ready for subsequent
workflow phases (Fig.~\ref{fig-workflow}, PyPotteryScan output block).

\subsubsection{Automated Content
Extraction}\label{automated-content-extraction}

\begin{figure}

\centering{

\includegraphics[width=0.8\linewidth,height=\textheight,keepaspectratio]{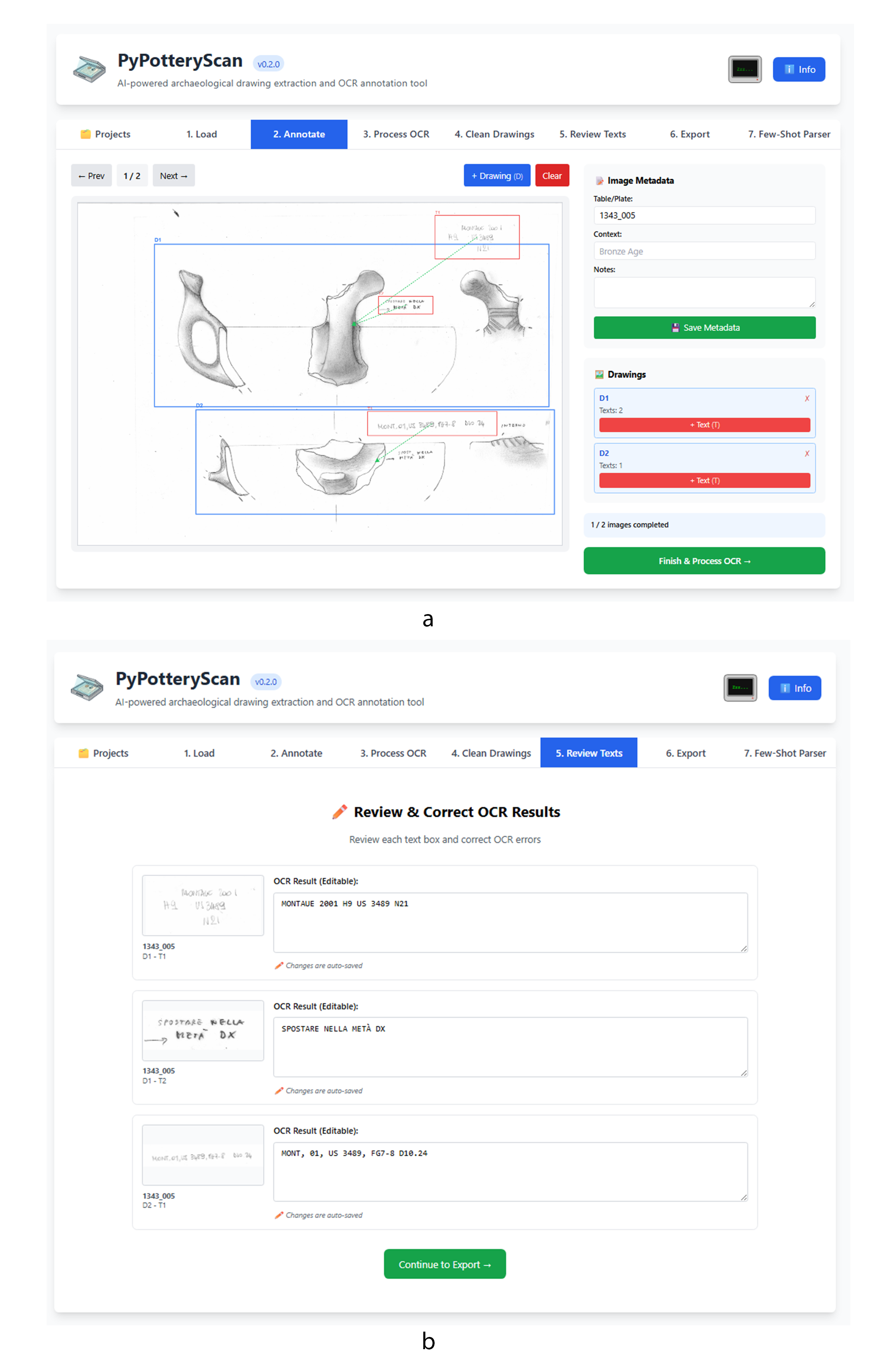}

}

\caption{\label{fig-scan_gui}Examples of the PyPotteryScan GUI v.2.0.0.
a) Annotated hand-drawn sheet. b) Manual OCR correction interface.}

\end{figure}

Text extraction presents greater complexity compared to image
extraction. In the vast majority of cases, textual information is
written in cursive handwriting, rendering traditional OCR systems
optimized for printed text inadequate. To address this challenge, olmOCR
was employed\footnote{olmOCR is fine-tuned on olmOCR-mix-0225, a
  training set of 260,000 pages drawn from over 100,000 crawled PDFs
  explicitly including handwritten text and poor-quality scans
  \cite{ref26}. Its multilingual capability is inherited from the
  underlying Qwen2-VL base model. The sheets from Montale were compiled
  by multiple individuals across different field seasons, ensuring
  handwriting variability in the evaluation sample.}, a 7B-parameter
open-source VLM specialized in handwriting recognition \cite{ref26}. The
model was quantized to 4-bit to significantly reduce size and
computational requirements, enabling execution on consumer hardware
\cite{ref27}. Given the inevitable presence of errors in automatic
recognition, a manual correction system was integrated into the GUI,
allowing users to quickly verify and correct any inaccuracies (Fig.
\ref{fig-scan_gui}b). The extracted text is then saved in XLSX format,
with each row paired to the corresponding drawing image via a unique
identifier.

The extracted textual information, although generally well-structured,
exhibits intrinsic variability that renders parsing based on fixed rules
(like regular expressions) inadequate. Factors such as diversity in
languages, stratigraphic context nomenclature, multiple conventions for
artifact inventories, and individual stylistic variations require a more
flexible and intelligent approach. For this reason, a custom LLM-based
parsing system was implemented, using the open-source Qwen3 7B-parameter
model \cite{ref28}. The system adopts a few-shot learning approach
\cite{ref29}: the user provides through the GUI a few examples of
correct parsing (typically 3--5 examples, see
Section~\ref{sec-results}), which are used as reference by the model to
correctly interpret the extracted textual information. This methodology
offers several advantages: the system easily adapts to different
documentation conventions without need for reprogramming, few examples
are sufficient to obtain accurate results, and it is executable on
consumer hardware thanks to the model's contained size.

The parsing output is saved in XLSX format, facilitating opening,
verification, and potential modification of data in common spreadsheet
software such as Microsoft Excel or Google Sheets as well as database
engines, thus ensuring maximum interoperability with existing workflows
(Fig.~\ref{fig-workflow}, PyPotteryScan output block).

\subsection{Vectorization and Editing with
PyPotteryTrace}\label{sec-trace}

\begin{figure}

\centering{

\includegraphics[width=0.75\linewidth,height=\textheight,keepaspectratio]{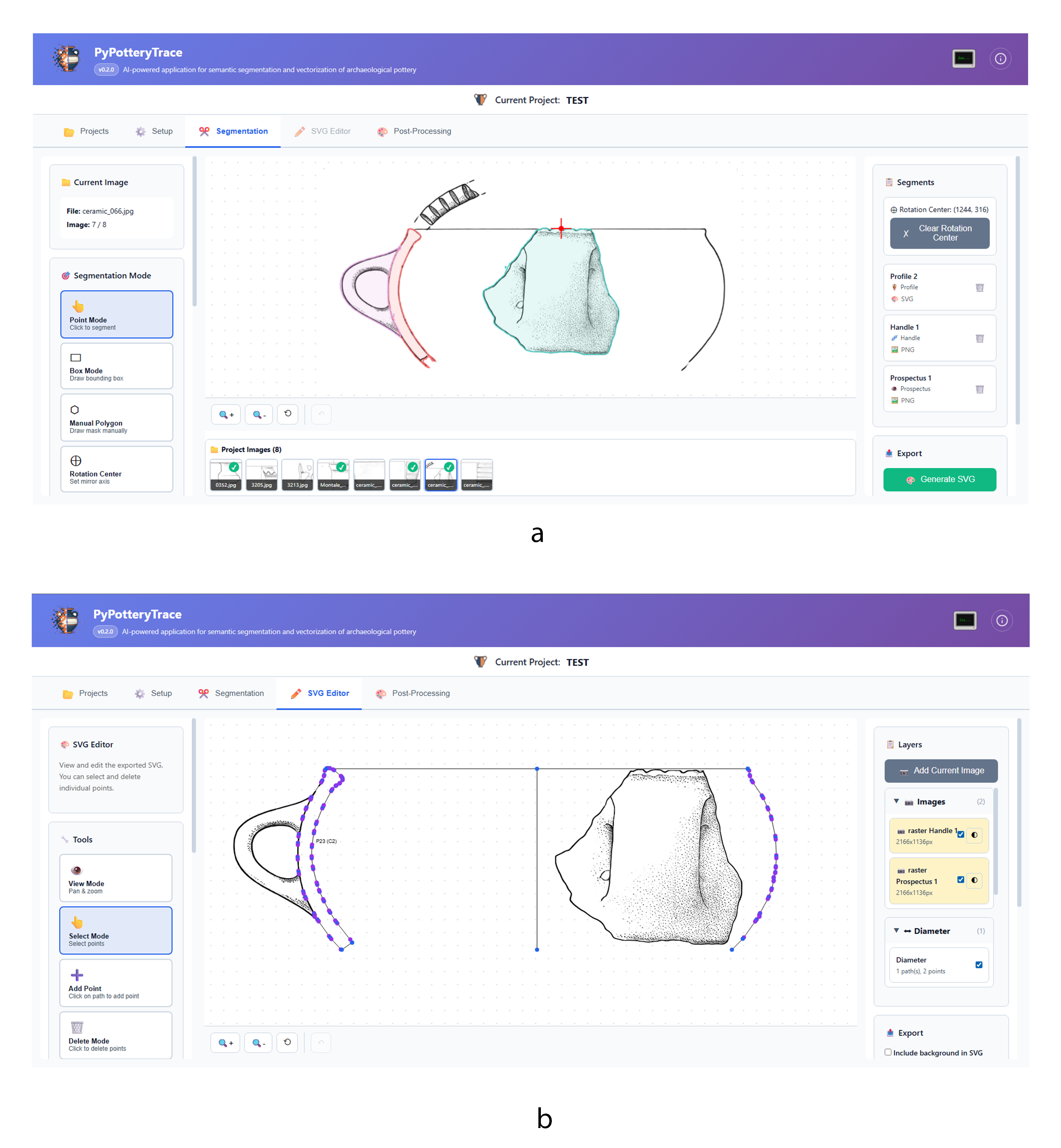}

}

\caption{\label{fig-trace_gui}Examples of the PyPotteryTrace GUI
v.2.0.0. a) Semantic segmentation. b) Built-in editing.}

\end{figure}

In the traditional documentation process, inking is performed using
vector graphics software, producing editable, scalable drawings. While
PyPotteryInk replicates the visual output of this process with
remarkable speed, it produces a raster image. PyPotteryTrace addresses
this limitation by converting raster outputs into structured vector
representations. Crucially, the motivation is not merely to enable
generic post-hoc editing, but to allow each semantic component of the
drawing --- profile, decoration, handle, etc. to be extracted, edited,
and exported independently. Solutions based on existing tools such as
Potrace (\url{https://potrace.sourceforge.net/}) or VTracer
(\url{https://github.com/visioncortex/vtracer}), as well as automatic
tracing tools integrated in vector graphics software, are limited
because: (1) they are unable to distinguish the various semantic
components of the drawing; (2) they represent stroke width by tracing
the outer contour of each line, producing closed filled polygons rather
than single-path centerlines. To manage this semantic complexity,
PyPotteryTrace adopts a component-based approach analogous to
PyPotteryScan, in which the user selects the various components of the
drawing and processes them separately. Users interactively segment the
drawing into specific archaeological classes (e.g., Profile,
Application, Handle, Prospectus, Decoration, Running Element, Detail)
assisted by the Segment-Anything 2 (SAM 2) model \cite{ref30}. This
interaction not only facilitates immediate vectorization but also
generates a labeled dataset of coordinate selection for future
fine-tuning (Fig. \ref{fig-trace_gui}a).

\begin{figure}

\centering{

\includegraphics[width=0.9\linewidth,height=\textheight,keepaspectratio]{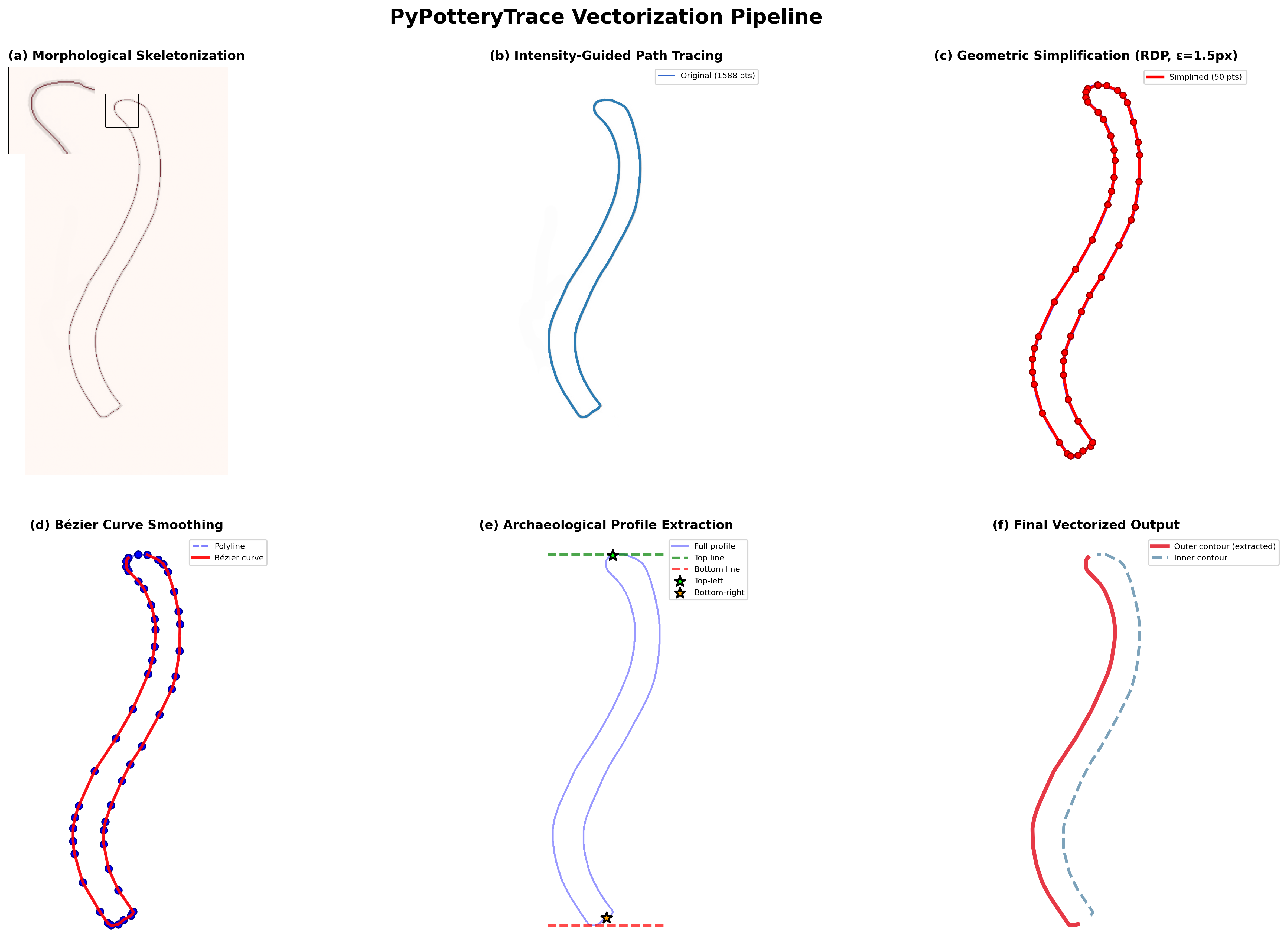}

}

\caption{\label{fig-vect_pipeline}Vectorization pipeline applied to a
ceramic profile. Each step is described in the text.}

\end{figure}

For each segmented component, the system applies a dedicated
vectorization pipeline. This process uses morphological and topological
transformations to convert the raster input into a semantically-aware
vector representation. The vectorization pipeline consists of five key
stages:

\begin{enumerate}
\def\labelenumi{\arabic{enumi}.}
\item
  \textbf{Morphological skeletonization}: The binary regions of the
  drawing are reduced to unit-width curves using morphological
  skeletonization, effectively identifying the ``medial axis'' of each
  line while preserving topological structure (Fig.
  \ref{fig-vect_pipeline}a).
\item
  \textbf{Intensity-guided path tracing}: PyPotteryTrace implements an
  intensity-guided tracing algorithm. It traverses the skeleton from
  endpoints, handling junctions by selecting directions that maintain
  angular continuity and follow darker pixel paths in the original
  image. This filters out spurious branches and noise during the tracing
  phase itself (Fig. \ref{fig-vect_pipeline}b).
\item
  \textbf{Geometric simplification}: The resulting raw paths are
  simplified using the Ramer--Douglas--Peucker (RDP) algorithm
  \cite{ref31}. This recursive method reduces the point count by
  retaining only those points necessary to represent the curve within a
  specified tolerance, significantly optimizing vector size (Fig.
  \ref{fig-vect_pipeline}c).
\item
  \textbf{Smoothing with Bézier curves}: To achieve the smooth
  ``archaeological line quality'', simplified paths are converted into
  cubic Bézier splines. Control points are calculated using the
  Catmull--Rom method \cite{ref32}, ensuring continuity (smooth
  tangents) between segments while passing exactly through the original
  points (Fig. \ref{fig-vect_pipeline}d).
\item
  \textbf{Archaeological profile extraction}: For ceramic profiles,
  PyPotteryTrace implements a specialized algorithm that extracts the
  vessel's external left contour, exploiting the radial symmetry
  characteristic of archaeological profile drawings (Fig.
  \ref{fig-vect_pipeline}e--f).
\end{enumerate}

Additional details can be found in the Supplementary materials --
Section S2.

The module includes also a manual vectorization mode (if automatic
vectorization fails), a comprehensive GUI for Bézier curve editing and a
batch export mode supporting standard formats (SVG, JPG, PNG), ensuring
full compatibility with vector software like Adobe Illustrator or
Inkscape (Fig.~\ref{fig-workflow}, PyPotteryTrace output block; Fig.
\ref{fig-trace_gui}b).

\subsection{Layout and Publication with
PyPotteryLayout}\label{layout-and-publication-with-pypotterylayout}

PyPotteryLayout automates the final phase of the archaeological
documentation workflow, traditionally performed in desktop publishing
software. The system offers two complementary layout strategies to
handle the diverse needs of publication: (1) Adaptive Grid Layout: this
algorithm positions artifacts within a user-defined matrix (e.g., a 2x6
grid). (2) Optimized Layout: the system utilizes a 2D bin packing
algorithm (\emph{rectpack} -- \url{https://github.com/secnot/rectpack})
for maximizing space efficiency. This approach treats the page as a
container and artifacts as rigid items, optimizing their arrangement to
minimize wasted white space.

The system enriches these layouts with a hierarchical sorting engine
(e.g., primary sort by Stratigraphic Unit, secondary sort by Inventory
Number), automatic metric scale calibration, and metadata-driven caption
generation (Supplementary materials, Fig. S3).

Additional details can be found in the Supplementary materials --
Section S3.

\subsection{Performance Metrics Definition and Usability Study
Design}\label{performance-metrics-definition-and-usability-study-design}

To evaluate the effectiveness of the proposed human-in-the-loop
workflow, a set of quantitative metrics is defined and measured on the
sample. The evaluation focuses on several critical aspects: processing
time efficiency, handwriting recognition accuracy, structured parsing
reliability, vectorization quality, perceived time saving and software
usability. For details about metrics and evaluation procedures, see
Supplementary materials -- Sections S1 and S4. The inking phase, as
previously validated \cite{ref20}, is not re-evaluated in this study.

\subsection{Implementation}\label{implementation}

The GUI is developed using Flask/HTML for the front-end and Python for
the back-end. This approach ensures cross-platform compatibility,
usability and responsive design. All technical details can be found in
the GitHub repository (\url{https://github.com/lrncrd/PyPottery}) and
the dedicated Wiki documentation
(\url{https://lrncrd.github.io/PyPottery/}). To ensure user-friendliness
a graphical installer is also available for Windows and macOS
(\url{https://lrncrd.github.io/PyPottery/suite_installation.html}). The
proposed framework operates entirely locally, without reliance on
external APIs or proprietary models, ensuring data privacy,
reproducibility, and long-term sustainability. All the experiments
presented in this work were conducted on consumer-grade hardware: an
Intel i7-12700K CPU, 48GB of RAM, and an NVIDIA RTX 3070ti GPU with 8GB
of VRAM. The entire pipeline is also tested on a laptop (Intel i7-10750H
CPU, 16GB of RAM, and an NVIDIA RTX 3060 GPU with 6GB of VRAM) and a
MacBook Air (M3 chip, 16GB RAM). All dependencies are listed in the
repositories.

\section{Results}\label{sec-results}

\subsection{Handwriting Recognition
Accuracy}\label{handwriting-recognition-accuracy}

OCR performance evaluation (Supplementary materials, Fig. S4)
demonstrates competitive accuracy with CER averaging \(6.2\%\)
(\(\pm 1.2\%\)) and WER at \(18.6\%\) (\(\pm 3.5\%\)). The CER of
\(\approx 6\%\) represents approximately 1 error per 17 characters. Such
an error density that, while requiring correction, is substantially
lower than the cost of complete manual transcription.

\subsection{Structured Parsing Performance and Few-Shot Learning
Dynamics}\label{structured-parsing-performance-and-few-shot-learning-dynamics}

Parsing evaluation (Fig.~\ref{fig-fewshot}, Table~\ref{tbl-fewshot})
reveals non-linear learning dynamics characteristic of in-context
learning in LLMs. The results expose three distinct performance regimes
across the few-shot continuum:

\begin{figure}

\centering{

\includegraphics[width=0.95\linewidth,height=\textheight,keepaspectratio]{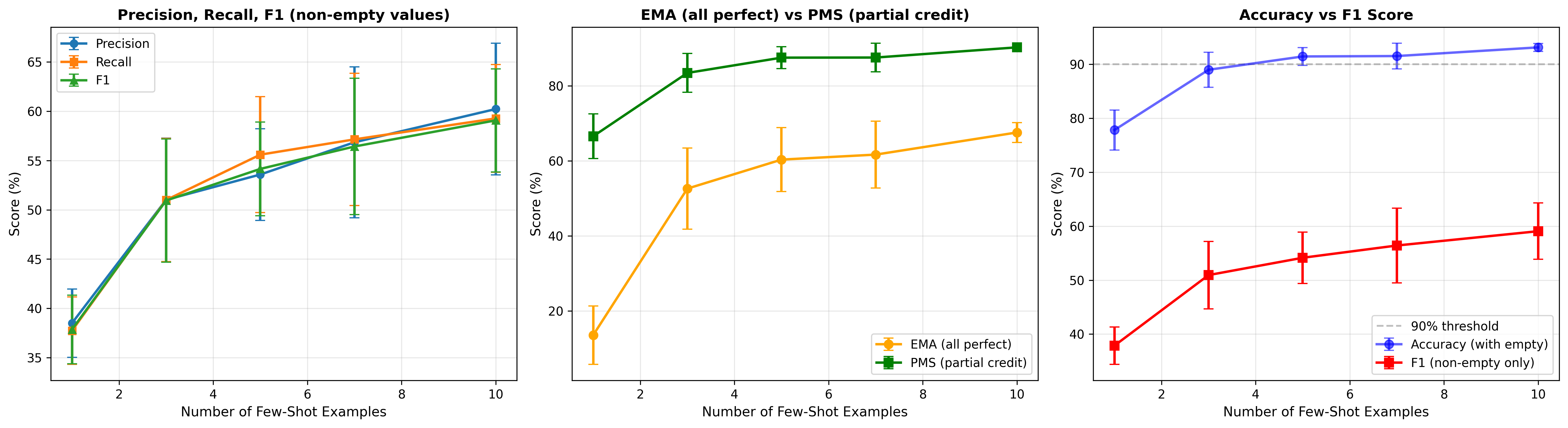}

}

\caption{\label{fig-fewshot}Few-shot learning efficiency across multiple
metrics. Left: Precision, Recall, F1 for non-empty field values. Center:
EMA (stringent all-fields-correct metric) vs.~PMS (partial credit
metric). Right: Accuracy (including empty-empty matches) vs.~F1
(non-empty only), with 90\% threshold marked. Error bars represent
\(\pm 1\) SD across 10 random example selections.}

\end{figure}%

\begin{table}

\caption{\label{tbl-fewshot}Few-shot learning performance summary. Metrics reported as mean $\pm$ standard deviation across 10 random samplings of training examples for each configuration.}

\centering{

[H]
\centering

\small
\begin{tabular}{c c c c c c c}
\toprule
\textbf{Num examples} & \textbf{Accuracy} & \textbf{Precision} & \textbf{Recall} & \textbf{F1} & \textbf{EMA} & \textbf{PMS} \\
\midrule
1  & 77.82 $\pm$ 3.72 & 38.50 $\pm$ 3.46 & 37.74 $\pm$ 3.42 & 37.86 $\pm$ 3.47 & 13.56 $\pm$ 7.80  & 66.59 $\pm$ 5.96 \\
3  & 88.99 $\pm$ 3.26 & 51.01 $\pm$ 6.27 & 50.99 $\pm$ 6.25 & 50.94 $\pm$ 6.25 & 52.63 $\pm$ 10.79 & 83.48 $\pm$ 5.19 \\
5  & 91.45 $\pm$ 1.64 & 53.58 $\pm$ 4.65 & 55.59 $\pm$ 5.87 & 54.15 $\pm$ 4.75 & 60.36 $\pm$ 8.52  & 87.55 $\pm$ 2.89 \\
7  & 91.53 $\pm$ 2.38 & 56.85 $\pm$ 7.65 & 57.14 $\pm$ 6.70 & 56.42 $\pm$ 6.92 & 61.70 $\pm$ 8.89  & 87.58 $\pm$ 3.82 \\
10 & 93.13 $\pm$ 0.73 & 60.22 $\pm$ 6.68 & 59.27 $\pm$ 5.45 & 59.07 $\pm$ 5.22 & 67.60 $\pm$ 2.65  & 90.30 $\pm$ 1.08 \\
\bottomrule
\end{tabular}

}

\end{table}%

The most dramatic improvement occurs between 1 and 3 examples, where EMA
(Exact Match Accuracy: proportion of records where all fields are
simultaneously correct) increases from 13.56\% to 52.63\%. Detailed
analysis (see Supplementary materials -- Section S1.2) reveals that
while 3 examples suffice for basic competence, increasing to 10 examples
primarily serves to reduce variance (\(\pm\) 2.65 SD). From a workflow
perspective, this reliability transforms the task from data creation to
rapid validation, with high PMS (Partial Match Score: average proportion
of correctly parsed fields per record, \(\approx\) 90\%) indicating that
most errors are isolated and easily correctable.

\subsection{Processing Time and Workflow
Efficiency}\label{processing-time-and-workflow-efficiency}

All processing times reported in this section were measured by the
author during a single end-to-end run of the complete pipeline on the
full dataset (50 sheets, 240 drawings). This evaluation is independent
from the usability study described in Section~\ref{sec-usability}, which
involved five domain experts working on a standardized subset of tasks.
Processing time analysis (Table~\ref{tbl-timing}) reveals the complete
workflow required approximately 103 minutes (1h 43min) for 50 sheets
containing 240 drawings, yielding a throughput of 140 drawings/hour
(2.33 drawings/minute). This aggregate figure masks important
distributional characteristics:

\begin{table}

\caption{\label{tbl-timing}Task-level time breakdown for complete PyPottery pipeline processing 50 sheets (240 drawings). ``Hardware impact'' indicates GPU dependency: High = GPU-accelerated (significant speedup), Medium = optional GPU acceleration, Low/Irrelevant = CPU-only operations (see Section 3, Implementation).}

\centering{

[H]
\centering

\small
\begin{tabular}{L{2.6cm}L{3.4cm}c L{1.9cm} L{1.9cm}}
\toprule
\textbf{Library} & \textbf{Task} & \textbf{Time} & \textbf{Hardware impact} & \textbf{Task type} \\
\midrule
\multirow{6}{2.6cm}{PyPotteryScan} & Manual filing & 13' & Irrelevant & Manual \\
 & Image extraction & 1' 13'' & Low & Automated \\
 & OCR & 5' 42'' & High & Automated \\
 & Cleaning & 10' 14'' & Irrelevant & Manual \\
 & OCR correction & 10' 40'' & Irrelevant & Manual \\
 & Parsing & 4' 3'' & High & Automated \\
 & Parsing correction & 3' 20'' & Irrelevant & Manual \\
\midrule
PyPotteryInk & Inking & 28' & High & Automated \\
\midrule
\multirow{2}{2.6cm}{PyPotteryTrace} & Manual segmentation & 25' & Medium & Mixed \\
 & Exporting & 30'' & Low & Automated \\
\midrule
PyPotteryLayout & Table creation \& Exporting & 1' 30'' & Irrelevant & Automated \\
\bottomrule
\end{tabular}

}

\end{table}%

Manual interventions (filing, cleaning, OCR correction, parsing
verification) account for approximately 37 minutes (36\% of total time),
while fully automated processes consume approximately 41 minutes (40\%),
with mixed semi-automated tasks (segmentation with SAM 2 assistance)
requiring 25 minutes (24\%). The manual control and correction permits
to handle errors and ensure archaeological validity but increases total
processing time.

The manual filing (13') and cleaning (10'14'') phases are prime
candidates for further automation. The coordinate annotation data
already collected during manual filing can train object detection models
(e.g., YOLO, SAM) to automatically localize drawing boxes and text
regions in future sheets. Similarly, the cleaning task (currently
performed with manual erasure tools) could be automated through advanced
models (similarly to PyPotteryInk) trained to remove grid lines,
annotations, and artifacts while preserving drawing content.

\subsection{Vectorization Output Quality: Quantitative and Qualitative
Assessment}\label{vectorization-output-quality-quantitative-and-qualitative-assessment}

\begin{figure}

\centering{

\includegraphics[width=0.9\linewidth,height=\textheight,keepaspectratio]{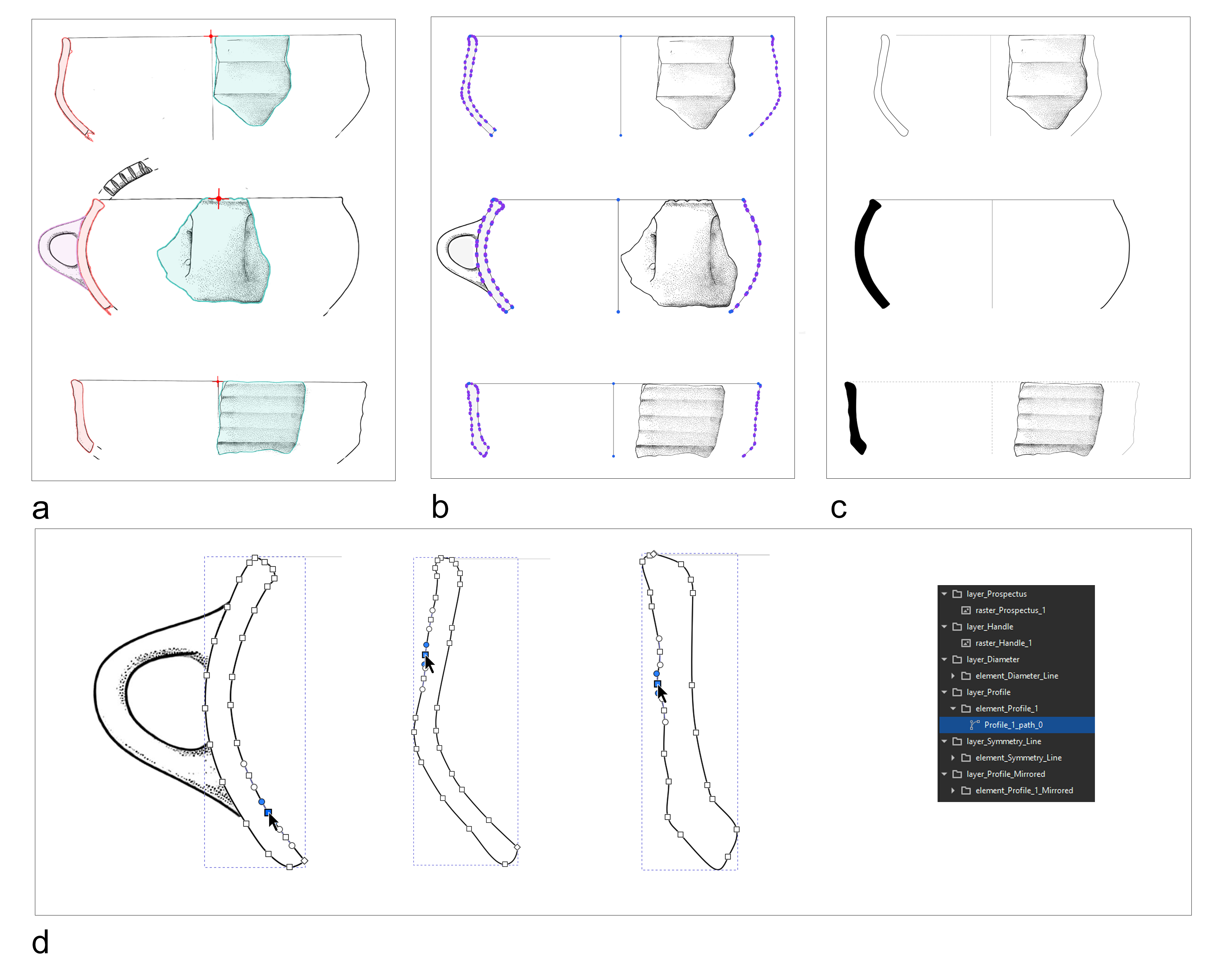}

}

\caption{\label{fig-trace_outputs}Representative outputs from
PyPotteryTrace showing automated vectorization of pottery profiles. a)
annotated image; b) vectorized profile (each point is connected with a
Bézier curve); c) post-processed vectors modified in vector-graphics
software; d) Bézier curves-styled profile and semantic vector
representation in Inkscape; the object tree (right) shows the semantic
layer structure of the exported SVG file, with independently editable
components (e.g., layer\_Profile, layer\_Handle, layer\_SymmetryLine,
layer\_Profile\_Mirrored).}

\end{figure}

The automated vectorization process required manual intervention for 21
records (8.75\% of the total). Failure cases fell into two main
categories: (1) pottery featuring intricate surface details such as
small applied clay stripes, where fine decorative elements were
difficult to isolate (e.g., Fig.~\ref{fig-trace_outputs},
\emph{ceramic\_117}); (2) open vessel forms (e.g., bowls) with handles
whose lower attachment closely followed the outer profile, creating
overlapping contours that the extraction algorithm could not reliably
disentangle.

Comparison between manual and automatic vectorization outputs shows high
correspondence in terms of visual appearance (Supplementary materials,
Fig. S5). Supplementary materials, Table S1 reports -- for reference --
the Hausdorff distance \cite{ref33} between manual and automatic
vectorization outputs. See Supplementary materials -- Section S2.6 for
further details.

Fig.~\ref{fig-trace_outputs} provides qualitative illustration of
PyPotteryTrace outputs: semantic features are successfully identified
(Fig. \ref{fig-trace_outputs}a) and extracted as Bézier paths suitable
for mirroring around symmetry axes (Fig. \ref{fig-trace_outputs}b--d).
The resulting vectors maintain the ``archaeological line quality''
typical of manual traced vectorized drawings (clean, smooth, continuous
curves) and enable further vector editing in standard vector graphics
software (Fig. \ref{fig-trace_outputs}c).

\begin{figure}

\centering{

\includegraphics[width=0.55\linewidth,height=\textheight,keepaspectratio]{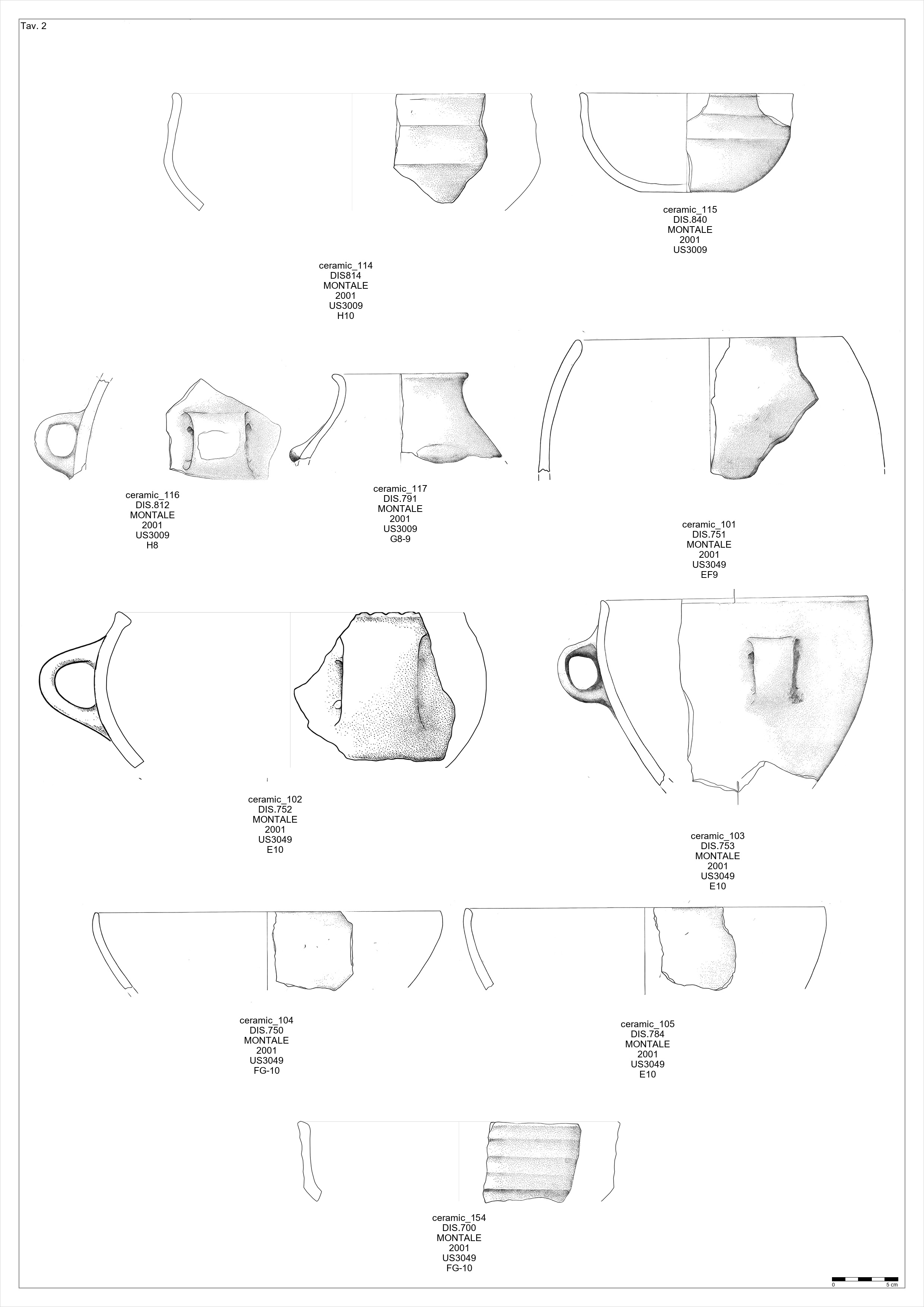}

}

\caption{\label{fig-layout_output}Representative outputs from
PyPotteryLayout showing automated publication layout generation.
Materials are organized using hierarchical sorting (primary:
stratigraphic unit; secondary: inventory number), with automatically
calibrated metric scale, metadata-derived captions, and optimized
spatial packing.}

\end{figure}

Fig.~\ref{fig-layout_output} shows PyPotteryLayout outputs.

The layout algorithm (in this case, a 3x5 grid) optimizes spatial
density while preserving archaeological logic, ensuring materials from
the same stratigraphic unit (for example) are grouped. Captions and
metric scales are automatically generated and positioned.

\subsection{Perceived Time Saving and User
Experience}\label{sec-usability}

A formal usability evaluation was conducted with five domain experts
from three major Italian institutions (Sapienza University of Rome,
University of Bologna, University of Naples ``Federico II''), averaging
13.4 years of archaeological experience (Appendix \ref{sec-appA}). The
sample size follows the established basis for formative usability
evaluation, five participants being sufficient to surface approximately
85\% of usability problems \cite{ref35}; it is not intended as a basis
for summative, multi-site evaluation (see
Section~\ref{sec-limitations}). All participants used Adobe Illustrator
as their primary documentation software, and none had previously used
AI-powered tools for archaeological work.

The full survey, detailed results and metric definitions are provided in
Supplementary materials -- Section S4 and Appendix \ref{sec-appA}.

The system demonstrated high effectiveness with an overall task
completion rate of 83.3\% across the six standardized tasks. PyPottery
achieved a mean System Usability Scale (SUS) score of 83.5 (SD = 11.26),
corresponding to Grade A on Brooke's standardized scale \cite{ref34}.
Participants unanimously perceived significant time savings across all
modules, with PyPotteryLayout achieving the highest ratings (5.0/5).
When asked to estimate processing time for 100 drawings, the median
perceived speedup was 40\(\times\) faster than traditional workflows
(mean: 61.9\(\times\); range: 17.5\(\times\) to 120\(\times\)), the
median being reported in preference to the mean given the small sample
and the wide dispersion of estimates. Inking/Vectorization was
identified as providing the greatest time savings by four of five
participants. Participants reported a median manual inking time of 15
minutes per drawing (range: 10--180), confirming that PyPottery's batch
processing represents a substantial efficiency gain.

Given the heterogeneity of the participant pool (4 to 25 years of
ceramic documentation experience), results were disaggregated by career
stage (Supplementary materials, Appendix \ref{sec-appA}). Early-career
participants (MA/MSc level, 4--5 years) returned identical SUS scores of
92.5, whereas post-doctoral participants (9--25 years) averaged 77.5 (SD
= 10.90, range 65.0--85.0). Across the sample, usability scores were
strongly and inversely related to documentation experience (Pearson
\(r = -0.97\) with years of ceramic experience; \(r = -0.96\) with years
of archaeological experience). Given \(n = 5\), these coefficients are
reported as descriptive of the present sample and are not intended to
support inference to a wider population.

Notably, this gradient does not extend to perceived efficiency:
archaeological experience is essentially uncorrelated with estimated
speedup (\(r = -0.003\)), and ceramic documentation experience only
weakly related to it (\(r = 0.12\)); the participant returning the
lowest SUS score (65.0) nonetheless estimated a 100\(\times\)
improvement over their traditional workflow. The two dimensions
therefore appear to be decoupled: experienced practitioners recognise
the efficiency gain as readily as early-career users, but encounter
greater friction in operating the interface.

This pattern is consistent with the task-level data. Six of the seven
recorded help requests concentrated in the vectorization task (T5),
which also recorded the lowest completion rate (1/5 complete, 4/5
partial); the single remaining request occurred in T1, which every
participant nonetheless completed. The difficulty is thus localised in
the semantic segmentation interface rather than distributed across the
suite, and plausibly reflects the cost of displacing long-established
Adobe Illustrator workflows --- all participants reported Illustrator as
their primary tool, and the three post-doctoral participants had each
processed over 1000 drawings within it. Rather than a general usability
deficit, the gradient identifies a specific design target: reducing the
interaction cost of PyPotteryTrace for users with entrenched alternative
workflows (see Section~\ref{sec-limitations}).

Interview analysis revealed that participants consistently praised
PyPottery's GUI quality and its alignment with archaeological workflows
(``Each step is well defined for archaeological purpose''). Crucially,
participants highlighted the system's independence, noting its potential
to ``completely replace traditional workflows'' without reliance on
expensive commercial software licenses. The most common difficulty cited
was the installation process (3/5 participants), while feature requests
centered on prospectus vectorization and improved dotting control in
inking. The vectorization task (PyPotteryTrace) presented the greatest
usability challenge, with participants reporting difficulties in the
semantic segmentation interface, suggesting this module may require
further refinement or better documentation.

\section{Discussion}\label{discussion}

Results will be discussed starting from the pipeline performance in
terms of time efficiency and quality of the outputs, then moving to the
wider implications of the work in computational archaeology.

\subsection{Performance and User
Validation}\label{performance-and-user-validation}

While the quantitative results are reported in detail in
Section~\ref{sec-results}, two interpretive points merit emphasis.
First, although manual and mixed tasks constitute approximately 60\% of
total workflow time, the majority of this involves \emph{reviewing}
algorithmic outputs rather than \emph{creating} content from scratch, a
fundamental shift in the nature of human labor. Second, the
human-in-the-loop design is not a limitation but a deliberate
requirement: stratigraphic and inventory data form the backbone of
archaeological records, and the time cost of manual verification is an
investment in data integrity. The usability results confirm that this
balance is well-calibrated: domain experts with no prior AI experience
could operate the system effectively, suggesting that interface
accessibility is a fundamental prerequisite for adoption.

\subsection{Redefining the Human-Machine Boundary in Archaeological
Interpretation}\label{redefining-the-human-machine-boundary-in-archaeological-interpretation}

The integration of AI into archaeological practice, as exemplified by
PyPottery, prompts fundamental epistemological reflection on the nature
of the documentary process itself. Automating mechanical and repetitive
post-production tasks is not about delegating archaeological
interpretation to machines. Rather, it constitutes a strategic
reallocation of cognitive labor that liberates researcher attention for
higher-value activities demanding irreplaceable human expertise: the
initial interpretive pottery drawing (\emph{for the time being}),
critical analysis of extracted data, and synthesis of findings into
coherent archaeological narratives.

However, this reallocation must be carefully calibrated to avoid
\emph{cognitive offloading}, the risk of delegating not merely physical
tasks but the intellectual processes of perceiving, deciding, and
interpreting to automated systems \cite{ref36}. In many digital
workflows, humans remain ``in the loop'' but are reduced to validators
of algorithmic outputs or data labelers for training sets, a phenomenon
characterized as \emph{heteromation} \cite{ref37}. PyPottery's design
explicitly resists this trajectory by automating only demonstrably
mechanical operations (image extraction, OCR transcription,
vectorization algorithms) while reserving all semantic and interpretive
judgments for the archaeologist. In doing so, the framework moves toward
what Huggett terms \emph{augmentation} rather than \emph{automation}
\cite{ref37} --- enhancing human capabilities rather than replacing
them. PyPottery thus fits within the Human-in-the-Loop paradigm
\cite{ref6}, where questions and results remain fundamentally
humanocentric.

PyPottery's modular design allows researchers to critically select which
steps to augment. If, for instance, the creation of publication plates
(automated by PyPotteryLayout) is viewed not as a repetitive task but as
a genuine ``thinking tool'' for understanding the ceramic assemblage,
the archaeologist can bypass this module and perform the layout
manually.

This redistribution challenges established notions of archaeological
skill and expertise. Proficiency in producing publication-quality ink
drawings or mastering vector graphics software, while valuable, does not
constitute the core of archaeological expertise. The discipline's
intellectual center lies in interpretation, material culture analysis,
and synthetic argumentation --- activities that remain fundamentally
human.

\subsection{Rethinking Archaeological Documentation Workflow
Design}\label{rethinking-archaeological-documentation-workflow-design}

As a concluding thought, PyPottery's development and deployment compels
broader reconsideration of archaeological documentation practices. The
availability of AI-powered automation necessitates strategic workflow
redesign that optimizes the allocation of human and machine labor:

\textbf{Principle 1: Delegate mechanical tasks to machines.} Repetitive,
rule-based operations (image cropping, text extraction, layout
generation) consume archaeologists' time without requiring specialized
domain knowledge. \emph{Augmentation} liberates cognitive resources for
interpretive work.

\textbf{Principle 2: Retain human control over archaeological
judgments.} Decisions about stratigraphic relationships, artifact
classification, and interpretive significance must remain under
archaeologist oversight. AI outputs should function as suggestions
requiring validation, not autonomous determinations. This aligns with
calls for explainable AI in archaeology \cite{ref6}, where algorithmic
decision-making processes must be transparent and archaeologically
interpretable.

\textbf{Principle 3: Design for upstream optimization.} Knowing that
automated systems will process field documentation --- whether OCR
parsing handwritten metadata or AI inking pencil drawings --- excavation
teams can adapt data collection practices to improve downstream
automation. Regarding the application of PyPottery:

\begin{itemize}
\item
  Creating clean, high-contrast drawings with consistent line weights
  facilitates automated inking and vectorization.
\item
  Using legible, consistent handwriting (even if slightly slower)
  significantly reduces OCR error rates.
\item
  Standardizing metadata terminology enhances parsing reliability.
\end{itemize}

\subsection{Limitations}\label{sec-limitations}

PyPottery's performance relies on standard archaeological documentation
quality (minimum 300 DPI scans, legible handwriting). Suboptimal inputs
do not break the workflow but increase the need for manual correction.
Regarding hardware, while the system is optimized for consumer-grade
equipment, it benefits from modern GPUs. However, the required
specifications remain well within the typical budget of individual
researchers or university departments: at the time of writing, a
consumer-grade laptop equipped with an entry-level dedicated GPU (e.g.,
NVIDIA RTX 5060) is available for approximately --1200.

The stratified usability results (Section~\ref{sec-usability}) point to
a further, related limitation: the interaction cost of the semantic
segmentation interface rises for users with established vector-graphics
workflows, and reducing it is a priority for the next development cycle.
Another significant limitation, also highlighted by test participants,
concerns the vectorization of complex vessel views (prospectus).
Currently, accurately vectorizing drawings with rich surface decorations
or extensive shading remains challenging. Further software development
is required to address this limitation, potentially involving the
development of domain-specific vectorization algorithms or specialized
computer vision models.

A further limitation concerns verification. All source code, model
weights, evaluation outputs and raw usability data are publicly
available, but the source drawings underlying the results reported in
Sections 4.1--4.4 cannot be released, as the material is unpublished
(Section 3.3). The results presented here can therefore be inspected and
audited, but not re-executed on identical inputs. This constraint is
intrinsic to the application domain rather than incidental to this
study: the suite is designed precisely for the legacy and unpublished
documentation that is least available for public benchmarking.

Finally, the evaluation is based on a single site and documentation
tradition. While 50 sheets containing 240 drawings constitute a
substantial sample for a feasibility study, generalization to different
sites, languages, scan qualities, and pottery traditions remains to be
formally assessed. Being open-source and freely available, PyPottery can
be readily applied and evaluated by any research group on their own
materials, and cross-site validation is identified as a priority for
future community-driven work. Early evidence of this applicability is
already emerging: at the time of writing, at least seven researchers
have begun applying PyPottery to material spanning different
chronologies, languages, and documentation traditions (Supplementary
materials, Appendix \ref{sec-appB}). These are cases known to the author
because the researchers made contact directly; the actual extent of use
is not known, as the software requires no registration. This uptake has
occurred prior to peer-reviewed publication, through disciplinary
circulation of the openly available repository; it nonetheless remains
an early stage of dissemination and does not substitute for the
systematic cross-site evaluation identified above.

\section{Conclusion and Future Work}\label{conclusion-and-future-work}

This paper has presented PyPottery as both a practical solution to
archaeological documentation bottlenecks and an exploration of AI's
evolving role in archaeological methodology. The framework's technical
achievements demonstrate the transformative potential of AI: processing
240 drawings in 103 minutes represents a substantial reduction in
processing time, consistent with the efficiency gains reported by
usability study participants.

Beyond these quantitative results, PyPottery addresses fundamental
questions about human-AI collaboration in archaeology. The framework
does not replace archaeological judgment but strategically redistributes
cognitive labor: machines handle mechanical tasks (transcription,
vectorization, layout generation), while archaeologists retain
interpretive authority over material analysis and data validation. This
redistribution challenges traditional skill hierarchies --- proficiency
in manual inking or database entry, while valuable, is distinct from
core archaeological expertise in material interpretation and synthetic
reasoning.

The transformative impact of AI on scientific research and archaeology
in particular can no longer be ignored. Harnessing this technology is
essential: an increasing number of AI-powered tools are emerging, many
adopting collaborative, community-driven approaches (e.g., CeraLoop:
\url{https://ceraloop.lad-sapienza.it}). PyPottery exemplifies how novel
AI-assisted tools can efficiently address ``old'' disciplinary
challenges with renewed efficiency.

Future development will pursue complementary directions across multiple
scales. Technical automation will leverage annotation data already
collected during manual phases: training object detection models (YOLO,
fine-tuned SAM 2) to automatically localize drawing boxes and text
regions, and developing image-to-image models to automate the cleaning
phase (removing grid lines and annotations). As annotated records and
semantic segmentation data accumulate through community usage,
parameter-efficient fine-tuning approaches such as LoRA/QLoRA could be
applied to Qwen3 and SAM 2 to further improve parsing accuracy and
segmentation quality for specific documentation traditions.
Concurrently, to lower the technical barrier to entry highlighted by the
usability study, the project will release comprehensive educational
materials, including video tutorials and simplified installation guides,
ensuring the tool is accessible to the widest possible archaeological
audience.

\section*{Declaration of Interest
statement}\label{declaration-of-interest-statement}
\addcontentsline{toc}{section}{Declaration of Interest statement}

The author declares that he has no known competing financial interests
or personal relationships that could have appeared to influence the work
reported in this paper.

\section*{Declaration of generative AI and AI-assisted technologies in
the manuscript preparation
process}\label{declaration-of-generative-ai-and-ai-assisted-technologies-in-the-manuscript-preparation-process}
\addcontentsline{toc}{section}{Declaration of generative AI and
AI-assisted technologies in the manuscript preparation process}

During the preparation of this work the author used Claude 4.5 Sonnet
and Gemini 3.0 Pro in order to improve readability, make corrections to
the code and create GUIs. After using this tool/service, the author
reviewed and edited the content as needed and takes full responsibility
for the content of the published article.

\section*{Fundings}\label{fundings}
\addcontentsline{toc}{section}{Fundings}

This research did not receive any specific grant from funding agencies
in the public, commercial, or not-for-profit sectors.

\section*{Data and Code Availability}\label{data-and-code-availability}
\addcontentsline{toc}{section}{Data and Code Availability}

PyPottery is released under an open licence at
\url{https://github.com/lrncrd/PyPottery}, including source code, model
weights, graphical installers for Windows and macOS, documentation and a
public issue tracker. The parsing evaluation outputs and the complete
few-shot prompt structure are available in the PyPotteryScan repository
(\texttt{/evaluation\_output}); the vectorization comparison outputs and
Hausdorff distance computations in the PyPotteryTrace repository
(\texttt{/comparison}); the complete raw usability study data, including
individual session records, in the PyPottery repository
(\texttt{usability\_test.zip}). The source drawings from the Terramara
di Montale cannot be released, as the material is unpublished and
remains under the control of the excavating institution (Section 3.3).

\section*{Acknowledgements}\label{acknowledgements}
\addcontentsline{toc}{section}{Acknowledgements}

The author would also like to thank Andrea Cardarelli (Fondazione
Sapienza) and Gianluca Pellacani (Museo Civico di Modena) for providing
the drawings used in the training dataset. The author further wishes to
thank S. Servoli, F.I. Debandi, A. Philippi, L. Schiavone, V. Parisi, E.
Fausti, and A. Di Renzoni for sharing their early application of
PyPottery to their respective research contexts. Thanks also to A.L.,
N.B., F.I.D., T.T., C.C. for participation in the usability study.

\clearpage
\appendix
\renewcommand{\thesection}{S\arabic{section}}
\setcounter{section}{0}
\renewcommand{\thetable}{S\arabic{table}}
\renewcommand{\thefigure}{S\arabic{figure}}
\setcounter{table}{0}
\setcounter{figure}{0}
\begin{center}
{\LARGE\bfseries Supplementary Materials}\\[0.3cm]
{\large PyPottery: an AI-powered end-to-end suite for pottery processing and publication}
\end{center}
\vspace{0.5cm}

\begin{figure}

\centering{

\includegraphics[width=0.8\linewidth,height=\textheight,keepaspectratio]{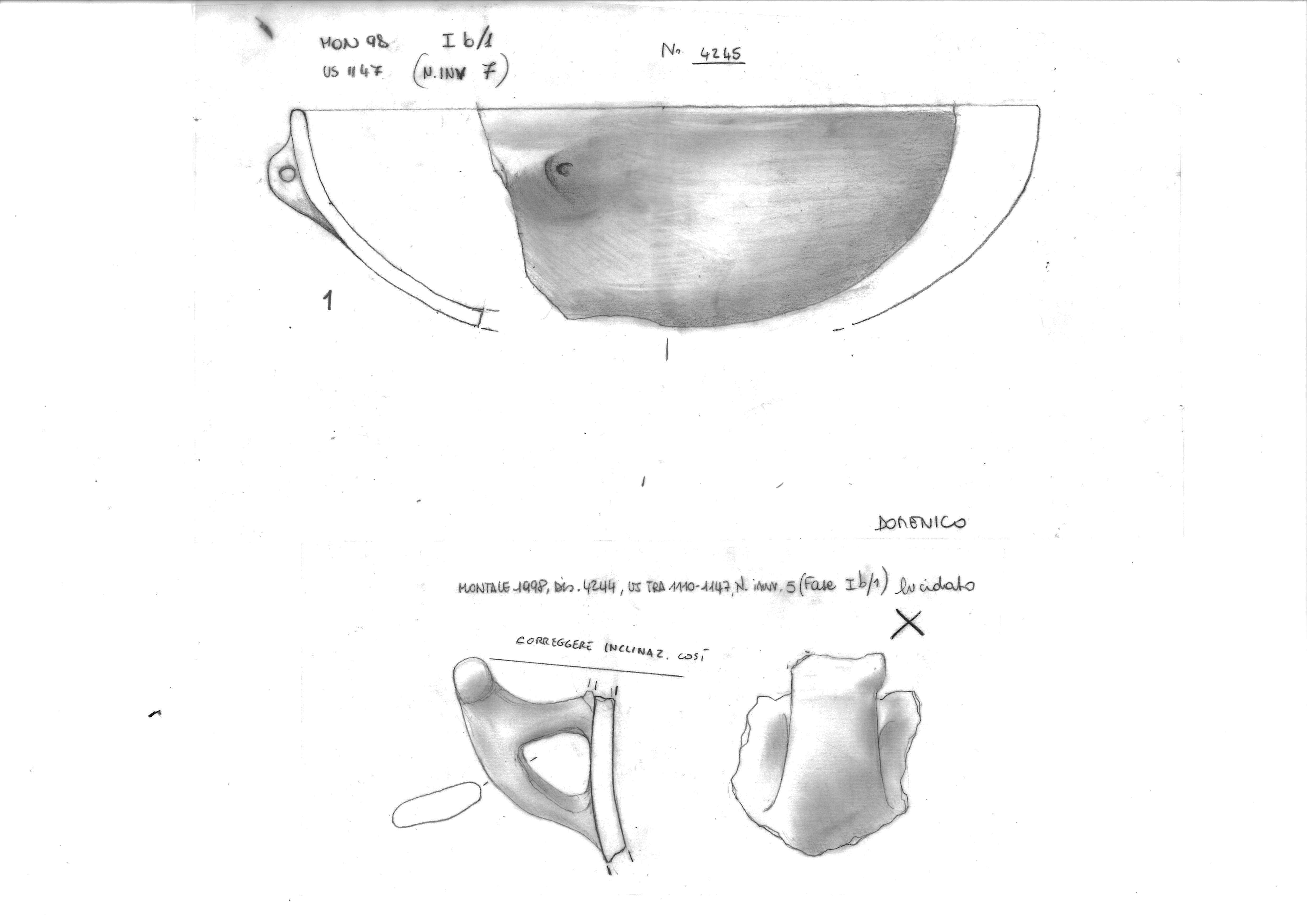}

}

\caption{\label{fig-S1}An example of a hand-drawn sheet from Montale.
The sheet contains pottery drawings and textual data including
stratigraphic information, inventory numbers and comments.}

\end{figure}%

\begin{figure}

\centering{

\includegraphics[width=0.7\linewidth,height=\textheight,keepaspectratio]{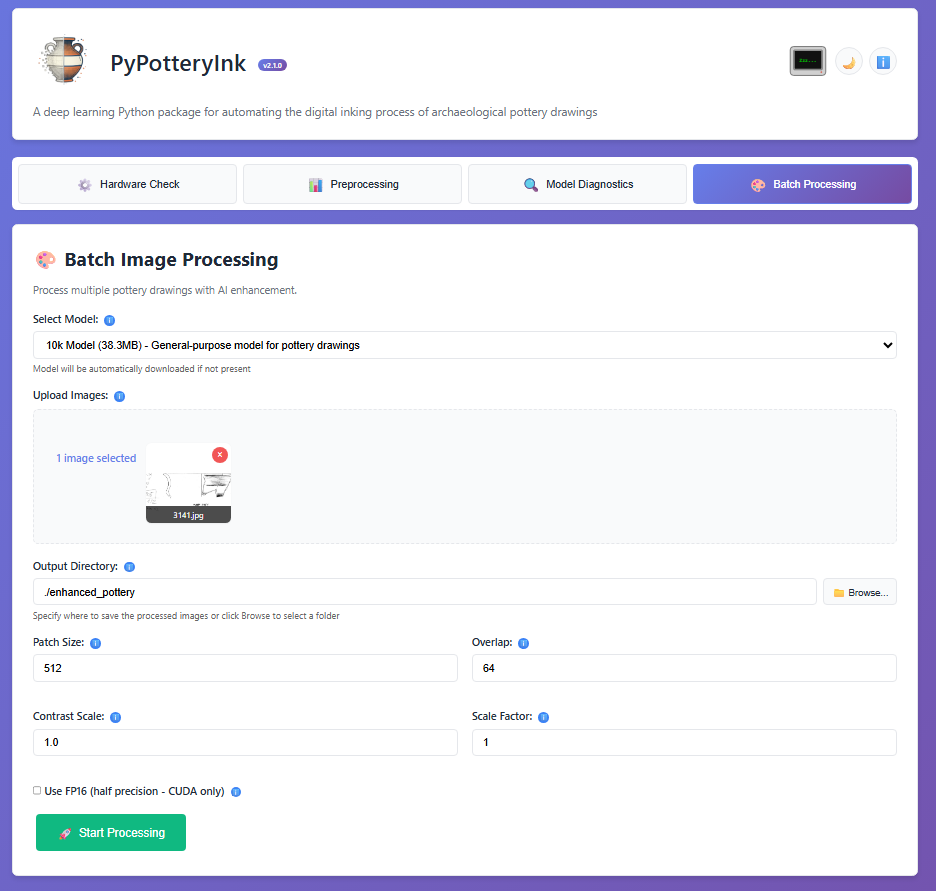}

}

\caption{\label{fig-S2}Example of the PyPotteryInk GUI v.2.1.0.}

\end{figure}%

\begin{figure}

\centering{

\includegraphics[width=0.7\linewidth,height=\textheight,keepaspectratio]{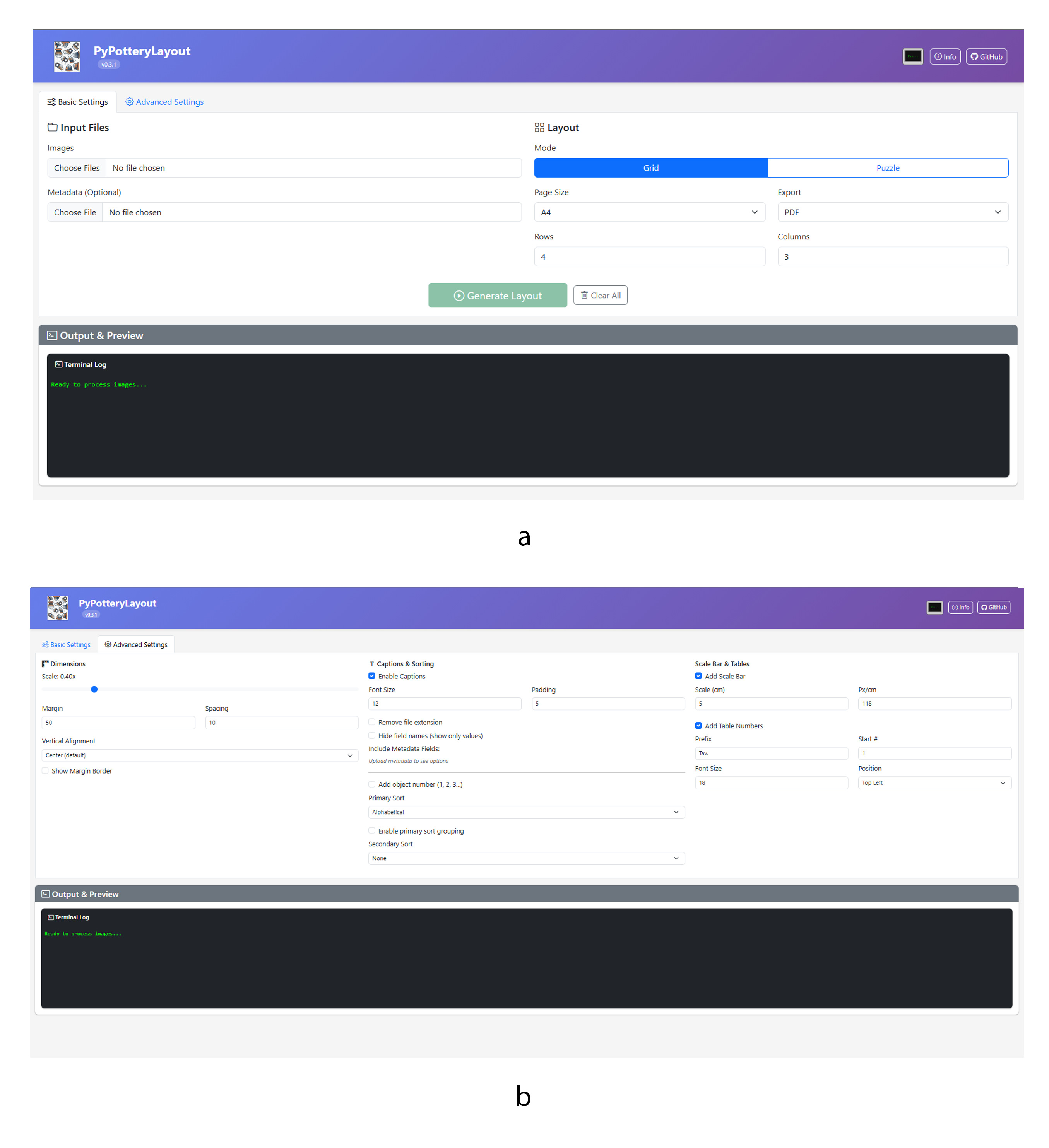}

}

\caption{\label{fig-S3}Example of the PyPotteryLayout GUI v.0.3.1. 1)
Basic settings 2) Advanced settings.}

\end{figure}%

\begin{figure}

\centering{

\includegraphics[width=0.5\linewidth,height=\textheight,keepaspectratio]{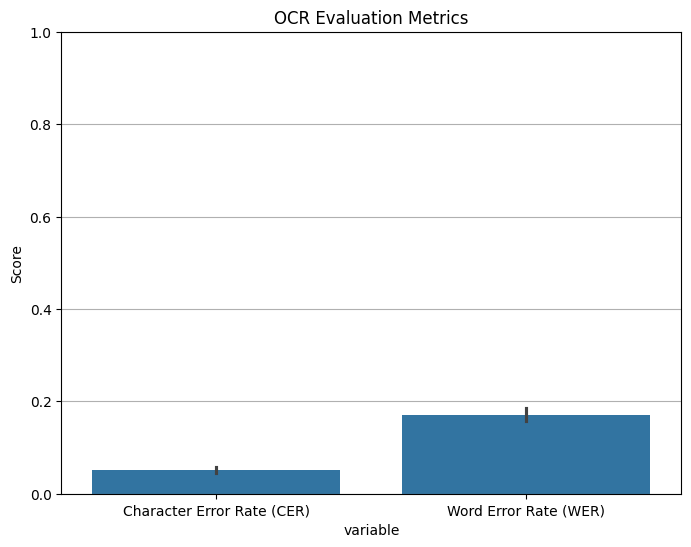}

}

\caption{\label{fig-S4}Handwriting recognition performance across
different metadata fields. Error bars represent \(\pm 1\) standard
deviation across 50 random iterations of 15-record samples.}

\end{figure}%

\begin{table}

\caption{\label{tbl-S1}Hausdorff distance between manual and automatic vectorization outputs.}

\centering{

[H]
\centering

\begin{tabular}{lc}
\toprule
\textbf{File} & \textbf{Hausdorff} \\
\midrule
0352.svg & 0.0266 \\
3205.svg & 0.0197 \\
3213.svg & 0.0211 \\
ceramic\_001.svg & 0.0123 \\
ceramic\_046.svg & 0.0086 \\
ceramic\_066.svg & 0.0254 \\
ceramic\_071.svg & 0.0211 \\
Montale\_109.svg & 0.0139 \\
\bottomrule
\end{tabular}

}

\end{table}%

\begin{figure}

\centering{

\includegraphics[width=0.8\linewidth,height=\textheight,keepaspectratio]{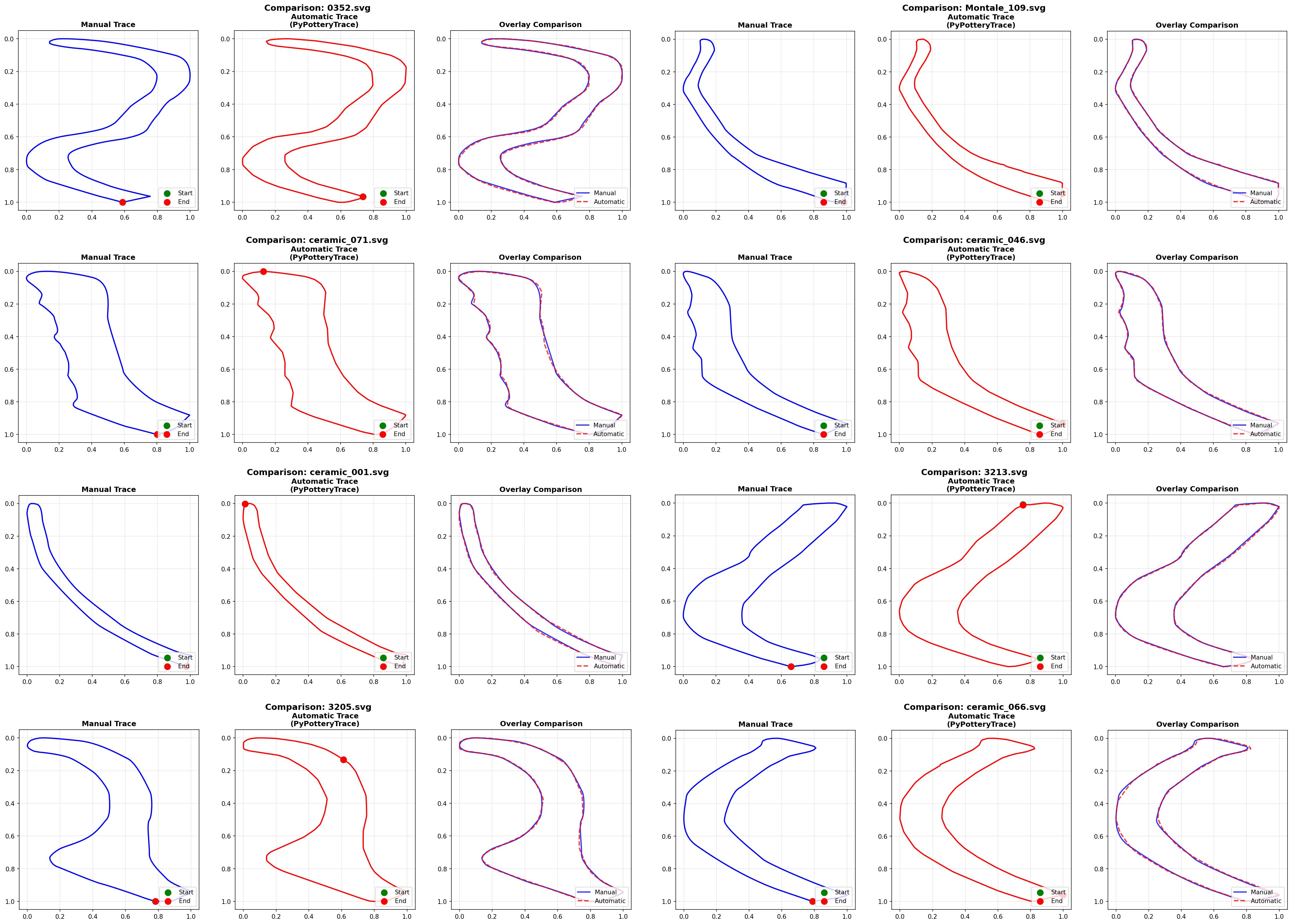}

}

\caption{\label{fig-S5}Comparison between manual and automatic
vectorization outputs.}

\end{figure}%

\section{Performance Metrics Definition and Usability Study
Design}\label{performance-metrics-definition-and-usability-study-design-1}

\subsection{Results Evaluation
Procedures}\label{results-evaluation-procedures}

\subsubsection{Processing Time and
Throughput}\label{processing-time-and-throughput}

Processing time is measured for each phase of the pipeline to identify
bottlenecks and estimate overall workflow efficiency. For a given sheet
\(S\) containing \(n\) artifacts:

\begin{itemize}
\item
  \textbf{Throughput} \(\Theta = \frac{1}{T_{artifact}}\): number of
  artifacts processed per unit time (artifacts/hour)
\item
  \textbf{Time for task} \(T_{task}\): time required to complete a
  specific task
\end{itemize}

\subsubsection{Handwriting Recognition
Accuracy}\label{handwriting-recognition-accuracy-1}

OCR performance is evaluated using standard metrics from the document
analysis literature. All 50 sheets were manually transcribed to create
ground truth annotations.

\textbf{Character Error Rate (CER)}: CER measures the minimum number of
character-level edits (insertions, deletions, substitutions) required to
transform the predicted text into the ground truth, normalized by the
ground truth length
(\url{https://huggingface.co/learn/audio-course/chapter5/evaluation\#character-error-rate}):

\begin{equation}
\text{CER} = \frac{S + D + I}{N} \cdot 100\%
\end{equation}

Where:

\begin{itemize}
\item
  \(S\): Substitutions (Number of incorrect characters recognized)
\item
  \(D\): Deletions (Number of characters present in the reference but
  missing in the hypothesis)
\item
  \(I\): Insertions (Number of extra characters present in the
  hypothesis but missing in the reference)
\item
  \(N\): Total number of characters in the reference sentence.
\end{itemize}

\textbf{Word Error Rate (WER)}: WER applies the same principle at the
word level, treating each whitespace-delimited token as an atomic unit
(\url{https://huggingface.co/learn/audio-course/chapter5/evaluation\#word-error-rate}):

\begin{equation}
\text{WER} = \frac{S + D + I}{N} \cdot 100\%
\end{equation}

Where: \(S\): Substitutions (Number of incorrect words recognized);
\(D\): Deletions (Number of words present in the reference but missing
in the hypothesis); \(I\): Insertions (Number of extra words present in
the hypothesis but missing in the reference); \(N\): Total number of
words in the reference sentence.

\subsubsection{Structured Parsing
Accuracy}\label{structured-parsing-accuracy}

LLM-based parsing performance was evaluated by measuring the model's
ability to correctly assign extracted textual information to predefined
metadata fields. The evaluation employed a ground truth dataset of
manually validated records, with each record containing \(k\) metadata
fields \(\mathcal{F} = \{f_1, f_2, \ldots, f_k\}\) (e.g., Stratigraphic
Unit, Inventory ID, Excavation Year, Site, Area, Cut, Sector, Phase,
Notes).

To ensure robust comparison and account for formatting inconsistencies
inherent in handwritten text, all values were normalized prior to metric
calculation. The normalization function \(\mathcal{N}(v)\) removes
spaces and non-alphanumeric characters, converting to lowercase:

\begin{equation}
\mathcal{N}(v) = \text{lowercase}\left( \text{remove\_non\_alphanumeric}(v) \right)
\end{equation}

This allows matches such as ``US 1147'' \(\equiv\) ``US1147'' and ``DIS.
N. 7'' \(\equiv\) ``DISN7'', while maintaining sensitivity to semantic
differences (e.g., ``US 1147'' \(\neq\) ``US 1148'').

\subsubsection{Field-Level Classification
Metrics}\label{field-level-classification-metrics}

For each field \(f_j\), parsing performance was evaluated using standard
classification metrics. Let \(n\) be the total number of test records.
For each record \(i\), we define:

\begin{itemize}
\item
  \(gt_{i,j}\): ground truth value for field \(f_j\) in record \(i\)
\item
  \(pred_{i,j}\): predicted value for field \(f_j\) in record \(i\)
\item
  A field is considered non-empty if \(\mathcal{N}(v) \neq \varnothing\)
  (after normalization)
\end{itemize}

\textbf{True Positives (TP)}: Non-empty ground truth correctly predicted
as non-empty with matching value:

\begin{equation}
TP_{j} = \left| \{ i : \mathcal{N}(gt_{i,j}) \neq \varnothing \land \mathcal{N}(pred_{i,j}) = \mathcal{N}(gt_{i,j}) \} \right|
\end{equation}

\textbf{False Positives (FP)}: Non-empty prediction when ground truth is
empty, or non-empty prediction with incorrect value:

\begin{equation}
\begin{split}
FP_{j} = & \left| \{ i : \mathcal{N}(gt_{i,j}) = \varnothing \land \mathcal{N}(pred_{i,j}) \neq \varnothing \} \right| \\
& + \left| \{ i : \mathcal{N}(gt_{i,j}) \neq \varnothing \land \mathcal{N}(pred_{i,j}) \neq \mathcal{N}(gt_{i,j}) \land \mathcal{N}(pred_{i,j}) \neq \varnothing \} \right|
\end{split}
\end{equation}

\textbf{False Negatives (FN)}: Non-empty ground truth predicted as empty
or with incorrect non-empty value:

\begin{equation}
\begin{split}
FN_{j} = & \left| \{ i : \mathcal{N}(gt_{i,j}) \neq \varnothing \land \mathcal{N}(pred_{i,j}) = \varnothing \} \right| \\
& + \left| \{ i : \mathcal{N}(gt_{i,j}) \neq \varnothing \land \mathcal{N}(pred_{i,j}) \neq \mathcal{N}(gt_{i,j}) \land \mathcal{N}(pred_{i,j}) \neq \varnothing \} \right|
\end{split}
\end{equation}

\textbf{True Negatives (TN)}: Empty ground truth correctly predicted as
empty:

\begin{equation}
TN_{j} = \left| \{ i : \mathcal{N}(gt_{i,j}) = \varnothing \land \mathcal{N}(pred_{i,j}) = \varnothing \} \right|
\end{equation}

From these, are derived:

\textbf{Accuracy}: Proportion of correct predictions (including
empty-empty matches):

\begin{equation}
\text{Acc}(f_j) = \frac{TP_j + TN_j}{n} \times 100\%
\end{equation}

\textbf{Precision}: Among non-empty predictions, proportion that are
correct:

\begin{equation}
\text{Precision}(f_j) = \frac{TP_j}{TP_j + FP_j} \times 100\%
\end{equation}

\textbf{Recall}: Among non-empty ground truths, proportion correctly
identified:

\begin{equation}
\text{Recall}(f_j) = \frac{TP_j}{TP_j + FN_j} \times 100\%
\end{equation}

\textbf{F1-Score}: Harmonic mean of precision and recall:

\begin{equation}
\text{F1}(f_j) = 2 \cdot \frac{\text{Precision}(f_j) \cdot \text{Recall}(f_j)}{\text{Precision}(f_j) + \text{Recall}(f_j)}
\end{equation}

\textbf{Macro-averaged metrics} across all fields provide overall
parsing quality:

\begin{equation}
\text{Metric}_{\text{macro}} = \frac{1}{k}\sum_{j=1}^{k}\text{Metric}(f_j)
\end{equation}

\subsubsection{Record-Level Metrics}\label{record-level-metrics}

Beyond field-level analysis, two record-level metrics are computed:

\textbf{Exact Match Accuracy (EMA)}: Proportion of records where all
fields are simultaneously correct:

\begin{equation}
\text{EMA} = \frac{1}{n}\sum_{i=1}^{n}\mathbb{1}\left[ \bigwedge_{j=1}^{k}\left( \mathcal{N}(pred_{i,j}) = \mathcal{N}(gt_{i,j}) \right) \right] \times 100\%
\end{equation}

This is a stringent metric requiring perfect parsing of the entire
record.

\textbf{Partial Match Score (PMS)}: Average proportion of correctly
parsed non-empty fields per record:

\begin{equation}
\text{PMS} = \frac{1}{n}\sum_{i=1}^{n}\frac{\sum_{j=1}^{k}\mathbb{1}\left[ \mathcal{N}(gt_{i,j}) \neq \varnothing \land \mathcal{N}(pred_{i,j}) = \mathcal{N}(gt_{i,j}) \right]}{\sum_{j=1}^{k}\mathbb{1}\left[ \mathcal{N}(gt_{i,j}) \neq \varnothing \right]} \times 100\%
\end{equation}

PMS provides partial credit for records with some correct fields,
offering a more nuanced view of parsing reliability than EMA.

\subsubsection{Few-Shot Learning
Efficiency}\label{few-shot-learning-efficiency}

The impact of few-shot learning was evaluated by varying the number of
in-context examples \(e \in \{1,3,5,7,10\}\) provided to the LLM. For
each configuration, the evaluation procedure was:

\begin{enumerate}
\def\labelenumi{\arabic{enumi}.}
\item
  Randomly sample \(e\) records as few-shot examples
\item
  Use remaining records as test set
\item
  Compute all metrics on test set
\item
  Repeat 10 times with different random samples to assess variance
\end{enumerate}

This yields metrics as a function of few-shot examples:

\begin{equation}
\text{Metric}(e) = \frac{1}{10}\sum_{r=1}^{10}\text{Metric}_r(e)
\end{equation}

where \(\text{Metric}_r(e)\) is the metric value for the \(r\)-th random
sampling with \(e\) examples. Standard deviation across repetitions
quantifies the sensitivity to example selection.

\subsection{Detailed Parsing Performance
Analysis}\label{detailed-parsing-performance-analysis}

The parsing evaluation reveals three distinct performance regimes. The
most dramatic improvement occurs in interval \(1 \rightarrow 3\): EMA
increases from 13.56\% to 52.63\%, while PMS jumps from 66.59\% to
83.48\%. This step indicates the model rapidly internalizes the parsing
schema. Beyond 3 examples, gains become modest but are characterized by
critical variance reduction: EMA standard deviation decreases from
\(\pm\) 10.79 to \(\pm\) 8.89. Further examples (\(7 \rightarrow 10\))
yield diminishing returns on absolute metrics but achieve substantial
variance stabilization (EMA SD drops to \(\pm\) 2.65), indicating the
model has saturated its capacity to learn the task structure.

The divergence between accuracy (\(\approx\) 93\%), F1 (\(\approx\)
59\%), and EMA (\(\approx\) 68\%) at 10 examples is notable. High
overall accuracy is partially inflated by true-negative matches
(legitimately empty fields). Precision and recall converge to
\(\sim\)60\%, indicating balanced performance without systematic bias.
EMA's ``all-or-nothing'' criterion (\(\approx\) 68\%) implies that while
one in three records contains an error, the high PMS (\(\approx\) 90\%)
reveals these errors are typically isolated. This validates the workflow
shift from data creation to verification: the system automates data
extraction, subject to verification for two-thirds of records, while the
remainder requires only targeted corrections. The variance reduction at
10 examples confirms that users can select training examples based on
convenience rather than statistical representativeness.

\subsection{Replicability}\label{replicability}

Results are available in
\url{https://github.com/lrncrd/PyPotteryScan/tree/main/evaluation_output}.
The complete prompt structure used for Qwen3 few-shot parsing, including
input format, output fields, example organization, and parsing
constraints, is documented in the same repository.

\section{Vectorization Pipeline}\label{vectorization-pipeline}

\subsection{Morphological
Skeletonization}\label{morphological-skeletonization}

The first stage is morphological skeletonization, an image processing
operation that reduces binary regions to unit-width curves while
preserving their topological structure. Intuitively, the skeleton
represents the ``medial axis'' of each drawn line --- the locus of
points equidistant from the region boundaries.

The implementation uses scikit-image's \texttt{skeletonize} function
(\url{https://scikit-image.org/docs/stable/auto_examples/edges/plot_skeleton.html\#skeletonize}),
which iteratively removes boundary pixels while maintaining 8-connected
continuity.

\subsection{Intensity-Guided Path
Tracing}\label{intensity-guided-path-tracing}

Once the skeleton is obtained, it must be converted into continuous
vector curves. Unlike traditional graph-based approaches that often
produce fragmented segments, PyPotteryTrace implements an
intensity-guided tracing algorithm that exploits information from the
original grayscale image.

The algorithm identifies two types of critical points on the skeleton:

\begin{itemize}
\item
  Endpoints: pixels with exactly one skeleton neighbor
\item
  Junctions: pixels with three or more skeleton neighbors
\end{itemize}

Path tracing proceeds by following the skeleton from endpoints,
selecting at each junction the direction that:

\begin{itemize}
\item
  Maintains directional continuity with the previous path segment
\item
  Follows darker lines in the original image (sampling up to 500 pixels
  ahead)
\item
  Avoids highly branched areas that typically indicate noise or spurious
  connections
\end{itemize}

This multi-criteria approach effectively filters out artifacts during
tracing rather than requiring post-processing cleanup.

\subsection{Geometric Simplification}\label{geometric-simplification}

The extracted paths, consisting of hundreds or thousands of pixel
coordinates, require simplification for efficient vector representation.
PyPotteryTrace employs the Ramer--Douglas--Peucker (RDP) algorithm
\cite{ref31}, a recursive method that preserves geometric features while
reducing point count.

Given a tolerance parameter \(\epsilon\), RDP identifies points that
deviate significantly from a straight-line approximation and retains
only those necessary to represent the curve within the specified
tolerance.

\subsection{Smoothing with Bézier
Curves}\label{smoothing-with-buxe9zier-curves}

To obtain visually natural curves, simplified paths are converted to
cubic Bézier splines using the Catmull--Rom method for calculating
control points
(\url{https://en.wikipedia.org/wiki/Catmull\%E2\%80\%93Rom_spline}).
Given a sequence of points, tangent vectors at each point are computed
from neighboring points:

\begin{equation}
\vec{t}_i = \tau \left( p_{i+1} - p_{i-1} \right)
\end{equation}

where \(\tau\) is a tension parameter (default 0.15). Control points are
then positioned at:

\begin{equation}
c_1 = p_i + \frac{\vec{t}_i}{3}, \qquad c_2 = p_{i+1} - \frac{\vec{t}_{i+1}}{3}
\end{equation}

This formulation guarantees \(C^1\) continuity (smooth tangents) between
adjacent curve segments while ensuring the curve passes exactly through
the original points.

\subsection{Archaeological Profile
Extraction}\label{archaeological-profile-extraction}

For ceramic profiles, PyPotteryTrace implements a specialized algorithm
that extracts the vessel's external left contour, exploiting the radial
symmetry characteristic of archaeological profile drawings.

The process:

\begin{enumerate}
\def\labelenumi{\arabic{enumi}.}
\item
  Identifies the topmost and bottommost points of the closed profile
  curve
\item
  Draws horizontal tangent lines at these extremes
\item
  Locates intersection points: the leftmost point on the top line and
  the rightmost point on the bottom line (to include flat bases)
\item
  Extracts the path segment between these points that lies furthest left
  (the external contour)
\item
  A configurable vertical tolerance (default 15 pixels) accommodates
  flat vessel bases, ensuring complete contour extraction regardless of
  base geometry.
\end{enumerate}

The mathematical formulation is available in the GitHub repository
(\url{https://github.com/lrncrd/PyPotteryTrace/blob/main/vectorization_pipeline_math.md}).

\subsection{Vectorization Quality Evaluation: Hausdorff
Distance}\label{vectorization-quality-evaluation-hausdorff-distance}

To quantitatively assess the quality of automatically generated vector
profiles, we compare PyPotteryTrace outputs against manually traced
ground truth using the Hausdorff distance, a well-established metric in
computational geometry for measuring the dissimilarity between two-point
sets \cite{ref33}.

Given two point sets \(A = \{a_1, a_2, \ldots, a_n\}\) and
\(B = \{b_1, b_2, \ldots, b_m\}\) representing sampled points along two
curves, the Hausdorff distance \(H(A,B)\) is defined as:

\begin{equation}
H(A,B) = \max\left\{ h(A,B), h(B,A) \right\}
\end{equation}

where \(h(A,B)\) is the directed Hausdorff distance from \(A\) to \(B\):

\begin{equation}
h(A,B) = \max_{a \in A} \min_{b \in B} \| a - b \|
\end{equation}

In other words, \(h(A,B)\) finds the point in \(A\) that is farthest
from any point in \(B\). The symmetric Hausdorff distance \(H(A,B)\)
takes the maximum of both directed distances, ensuring that the metric
is symmetric and captures the worst-case deviation in either direction.

Geometrically, it answers the question: \emph{``How far are the two
curves from being identical?''} by measuring the maximum deviation
between them.

For comparing pottery profile vectors:

\begin{enumerate}
\def\labelenumi{\arabic{enumi}.}
\item
  \textbf{Sampling}: Both manual and automatic SVG paths are uniformly
  sampled to obtain \(n\) points along each curve (typically \(n = 100\)
  points via arc-length parameterization).
\item
  \textbf{Normalization}: Points are normalized to \([0,1]\) range to
  ensure scale-invariant comparison:

  \begin{equation}
  \text{normalize}(P) = \frac{P - \min(P)}{\max(P) - \min(P)}
  \end{equation}
\item
  \textbf{Distance calculation}: The Hausdorff distance is computed
  using the Euclidean norm:

  \begin{equation}
  \| a - b \| = \sqrt{(a_x - b_x)^2 + (a_y - b_y)^2}
  \end{equation}
\end{enumerate}

The results are available in the GitHub repository
(\url{https://github.com/lrncrd/PyPotteryTrace/tree/main/comparison}).

\section{Layout Pipeline}\label{layout-pipeline}

\subsection{Layout Algorithms}\label{layout-algorithms}

The software employs two complementary approaches for organizing
artifacts on publication-ready pages.

\subsubsection*{Adaptive Grid Layout}\label{adaptive-grid-layout}
\addcontentsline{toc}{subsubsection}{Adaptive Grid Layout}

The system positions artifacts in a user-defined matrix of rows and
columns, implementing an adaptive allocation algorithm that handles
heterogeneous dimensional constraints. Each cell can host an artifact
scaled to fit within the available space. When an artifact's dimensions
exceed the available cell width, the system automatically allocates it
to a dedicated row spanning the full-page width. The artifact is then
horizontally centered within this row. This fallback mechanism ensures
that no artifact is excluded from the final layout, preserving the
integrity of the documentary dataset regardless of dimensional
heterogeneity.

\subsubsection*{Optimized Layout (2D Bin
Packing)}\label{optimized-layout-2d-bin-packing}
\addcontentsline{toc}{subsubsection}{Optimized Layout (2D Bin Packing)}

For maximum space efficiency, the system solves the two-dimensional bin
packing problem using the rectpack algorithm
(\url{https://github.com/secnot/rectpack}). Given a set of images with
varying dimensions and page constraints (including margins), the
algorithm minimizes unused white space while respecting all dimensional
boundaries.

\subsection{Two-Level Hierarchical Sorting
System}\label{two-level-hierarchical-sorting-system}

PyPotteryLayout implements a hierarchical sorting system that enables
artifact organization according to multiple structured criteria. Users
can define a primary sorting criterion and a secondary criterion applied
within each group defined by the primary.

\subsubsection*{Available Sorting
Criteria}\label{available-sorting-criteria}
\addcontentsline{toc}{subsubsection}{Available Sorting Criteria}

Primary Level:

\begin{itemize}
\item
  Alphabetical: lexicographic ordering by filename
\item
  Natural numeric: intelligent extraction and comparison of numeric
  sequences embedded in filenames (e.g., ``artifact\_2'' before
  ``artifact\_10'')
\item
  Archaeological metadata: sorting by any metadata field (context,
  stratigraphic unit, period, typology, etc.)
\item
  Random: controlled randomization for sampling or presentation variety
\end{itemize}

Secondary Level:

\begin{itemize}
\item
  Applied within each group defined by the primary criterion
\item
  Supports the same operators as the primary level
\item
  Permits fine refinement of the presentation sequence
\end{itemize}

Example application: Setting primary sorting by Stratigraphic Unit (SU)
and secondary sorting alphabetically produces layouts where all
materials from the same SU appear consecutively, internally sorted by
filename.

\subsection{Integration with Tabular
Metadata}\label{integration-with-tabular-metadata}

The system loads metadata from Excel (xlsx) or CSV files, where each row
associates a filename with multiple attribute fields. All available
fields automatically become usable as sorting criteria, enabling
customized workflows for different research contexts without code
modification.

You can find the mathematical formulation in the GitHub repository
(\url{https://github.com/lrncrd/PyPotteryLayout/blob/main/layout_pipeline_math.md}).

\section{Usability Study Design}\label{sec-S4}

Beyond computational performance metrics, a formal usability evaluation
was designed to assess whether domain experts can effectively use
PyPottery in practice and to validate perceived time savings against
empirical measurements.

\subsection{Study Rationale and
Design}\label{study-rationale-and-design}

The evaluation follows a within-subjects and task-based design, where
each participant completes standardized tasks across all PyPottery
modules while verbalizing their thought process. This approach was
chosen because: (1) within-subjects design maximizes statistical power
with limited participants by using each as their own control; (2)
task-based evaluation provides objective, comparable measures of
effectiveness and efficiency.

Sample size was set at five participants, following research
demonstrating that five users identify approximately 85\% of usability
problems \cite{ref35}, representing an optimal cost-benefit ratio for
formative usability studies. Participants were recruited based on three
criteria: an archaeological background (MA level or higher), at least
one year of experience with ceramic documentation, and basic computer
literacy. The latter was operationalized through a Technical Competency
Check designed to distinguish software usability issues from general
computer literacy gaps --- a potential confound that could otherwise
invalidate findings regarding PyPottery's learnability.

\subsection{Measures}\label{measures}

\textbf{Effectiveness} was measured through Task Completion Rate
(proportion of tasks successfully completed) and error counts.
\textbf{Efficiency} was measured through Time on Task for each of the
six standardized tasks covering all modules (sheet processing, OCR,
parsing, inking, vectorization, layout generation).

\textbf{Satisfaction} was assessed using the \textbf{System Usability
Scale (SUS)} \cite{ref34}, a validated 10-item questionnaire yielding
scores on a 0-100 scale with established benchmarks (scores above 68
indicate above-average usability).

\textbf{Perceived time savings} were captured through a custom
questionnaire asking participants to estimate time savings compared to
their traditional workflow for each module, using both Likert scales and
absolute time estimations. This subjective measure complements objective
timing data, as practitioners' perception of efficiency improvements is
crucial for adoption decisions.

\textbf{Qualitative insights} were gathered through semi-structured
interviews exploring overall impressions, comparison with traditional
workflows, and suggestions for improvement.

You can find the complete analysis and raw results in Appendix
\ref{sec-appA}.

\section{PyPottery Usability Study --- Results Analysis}\label{sec-appA}

This appendix contains the processed results of the PyPottery Usability
Study. The evaluation sessions were conducted individually with each
participant, either in person or remotely via videoconference, depending
on the participant's availability and location. Full methodological
details for each session are available in the raw PDF data in the GitHub
repository
(\url{https://github.com/lrncrd/PyPottery/blob/main/usability_test.zip}).

\FloatBarrier

\subsection{Participant Demographics}\label{participant-demographics}

\begin{table}[H]
\centering
\caption{Participant demographics.}
\small
\resizebox{\textwidth}{!}{%
\begin{tabular}{c c L{2.3cm} L{2.6cm} c c c L{2.6cm}}
\toprule
\textbf{ID} & \textbf{Age} & \textbf{Role} & \textbf{Institution} & \textbf{Arch. exp.\ (yrs)} & \textbf{Ceramic exp.\ (yrs)} & \textbf{Drawings processed} & \textbf{Software used} \\
\midrule
0 & 26--35 & MA/MSc & Sapienza University of Rome & 5 & 4 & 200--500 & Illustrator, Photoshop \\
1 & 26--35 & MA/MSc & University of Bologna & 6 & 5 & 500--1000 & Illustrator \\
2 & 46--55 & Post-doc & University of Bologna & 25 & 25 & $>$1000 & Illustrator \\
3 & 36--45 & Post-doc & University of Naples ``Federico II'' & 18 & 16 & $>$1000 & Illustrator, AutoCAD \\
4 & 26--35 & Post-doc & University of Naples ``Federico II'' & 13 & 9 & $>$1000 & Illustrator, CorelDRAW, AutoCAD \\
\bottomrule
\end{tabular}%
}
\end{table}

\begin{itemize}
\item
  Average archaeological experience: 13.4 years
\item
  Average ceramic documentation experience: 11.8 years
\item
  3/5 participants processed \(>\)1000 drawings in their career
\item
  All participants use Adobe Illustrator
\item
  None had previously used AI-powered tools for archaeological work
\item
  All passed the Technical Competency Check
\end{itemize}

\FloatBarrier

\subsection{System Usability Scale (SUS)
Results}\label{system-usability-scale-sus-results}

\FloatBarrier

\subsubsection*{Individual Scores}\label{individual-scores}
\addcontentsline{toc}{subsubsection}{Individual Scores}

\begin{table}[H]
\centering
\caption{Individual SUS item scores and totals.}
\small
\begin{tabular}{c ccccccccc c}
\toprule
\textbf{Part.} & \textbf{Q1} & \textbf{Q2} & \textbf{Q3} & \textbf{Q4} & \textbf{Q5} & \textbf{Q6} & \textbf{Q7} & \textbf{Q8} & \textbf{Q9} & \textbf{Q10} \quad \textbf{SUS} \\
\midrule
0 & 5 & 1 & 5 & 2 & 5 & 1 & 5 & 2 & 4 & 1 \quad \textbf{92.5} \\
1 & 5 & 1 & 5 & 2 & 5 & 1 & 5 & 2 & 5 & 2 \quad \textbf{92.5} \\
2 & 5 & 3 & 3 & 4 & 4 & 1 & 5 & 2 & 2 & 3 \quad \textbf{65.0} \\
3 & 5 & 1 & 5 & 4 & 5 & 1 & 5 & 4 & 5 & 2 \quad \textbf{82.5} \\
4 & 5 & 1 & 5 & 3 & 5 & 1 & 4 & 2 & 4 & 2 \quad \textbf{85.0} \\
\bottomrule
\end{tabular}
\end{table}

\begin{table}[H]
\centering
\caption{SUS summary statistics.}
\begin{tabular}{lll}
\toprule
\textbf{Metric} & \textbf{Value} & \textbf{Interpretation} \\
\midrule
Mean SUS Score & 83.5 & Grade: A (Excellent) \\
Standard Deviation & 11.26 & \\
Median & 85.0 & \\
Range & 65.0 -- 92.5 & \\
\bottomrule
\end{tabular}
\end{table}

\FloatBarrier

\subsection{Stratification by Participant
Profile}\label{stratification-by-participant-profile}

Because the participant pool spans a wide range of expertise (4--25
years of ceramic documentation experience), individual scores are
reported here against participant profile to enable disaggregated
inspection.

\begin{table}[H]
\centering
\caption{Individual profile, SUS score and estimated speedup.}
\small
\resizebox{\textwidth}{!}{%
\begin{tabular}{c l c c l c c}
\toprule
\textbf{ID} & \textbf{Role} & \textbf{Arch. exp.\ (yrs)} & \textbf{Ceramic exp.\ (yrs)} & \textbf{Drawings processed} & \textbf{SUS} & \textbf{Est.\ speedup} \\
\midrule
0 & MA/MSc & 5 & 4 & 200--500 & 92.5 & 120$\times$ \\
1 & MA/MSc & 6 & 5 & 500--1000 & 92.5 & 40$\times$ \\
2 & Post-doc & 25 & 25 & $>$1000 & 65.0 & 100$\times$ \\
3 & Post-doc & 18 & 16 & $>$1000 & 82.5 & 32$\times$ \\
4 & Post-doc & 13 & 9 & $>$1000 & 85.0 & 17.5$\times$ \\
\bottomrule
\end{tabular}%
}
\end{table}

\begin{table}[H]
\centering
\caption{Aggregates by career stage.}
\small
\resizebox{\textwidth}{!}{%
\begin{tabular}{l c c c c c c}
\toprule
\textbf{Career stage} & \textbf{n} & \textbf{Ceramic exp.} & \textbf{SUS (mean)} & \textbf{SUS (range)} & \textbf{Speedup (mean)} & \textbf{Speedup (range)} \\
\midrule
MA/MSc & 2 & 4--5 & 92.5 & 92.5--92.5 & 80.0$\times$ & 40$\times$--120$\times$ \\
Post-doc & 3 & 9--25 & 77.5 & 65.0--85.0 & 49.8$\times$ & 17.5$\times$--100$\times$ \\
\textbf{All} & \textbf{5} & \textbf{4--25} & \textbf{83.5} & \textbf{65.0--92.5} & \textbf{61.9$\times$} & \textbf{17.5$\times$--120$\times$} \\
\bottomrule
\end{tabular}%
}
\end{table}

Pearson correlations across the sample: ceramic experience \(\times\)
SUS, \(r = -0.97\); archaeological experience \(\times\) SUS,
\(r = -0.96\); archaeological experience \(\times\) estimated speedup,
\(r = -0.003\); ceramic experience \(\times\) estimated speedup,
\(r = 0.12\). Given \(n = 5\), these are reported as descriptive
statistics of the present sample and support no inferential claim.

\FloatBarrier

\subsection{Perceived Time Savings}\label{perceived-time-savings}

\FloatBarrier

\subsubsection*{By Module (Likert Scale
Aggregates)}\label{by-module-likert-scale-aggregates}
\addcontentsline{toc}{subsubsection}{By Module (Likert Scale
Aggregates)}

All responses indicate perceived time savings across all modules:

\begin{table}[H]
\centering
\caption{Perceived time savings by module.}
\small
\begin{tabular}{l L{4cm} L{6cm}}
\toprule
\textbf{Module} & \textbf{``Faster than traditional''} & \textbf{Time Savings Estimate} \\
\midrule
PyPotteryScan (T1--T3) & 4.8/5 (all Agree/Strongly Agree) & 4/5 = $>$75\% faster; 1/5 = 50--75\% faster \\
PyPotteryInk (T4) & 4.8/5 & 4/5 = $>$75\% faster; 1/5 = 25--50\% faster \\
PyPotteryTrace (T5) & 4.6/5 & 3/5 = $>$75\% faster; 1/5 = 50--75\% faster; 1/5 = 25--50\% faster \\
PyPotteryLayout (T6) & 5.0/5 (all Strongly Agree) & 5/5 = $>$75\% faster \\
\bottomrule
\end{tabular}
\end{table}

\FloatBarrier

\subsubsection*{Manual Inking Time
(Self-Reported)}\label{manual-inking-time-self-reported}
\addcontentsline{toc}{subsubsection}{Manual Inking Time (Self-Reported)}

Participants reported their typical time to ink one drawing manually:

\begin{table}[H]
\centering
\caption{Self-reported manual inking time per drawing.}
\begin{tabular}{cc}
\toprule
\textbf{Participant} & \textbf{Minutes per Drawing} \\
\midrule
0 & 40 \\
1 & 10--15 \\
2 & 30--180 \\
3 & 15 \\
4 & 13 \\
\bottomrule
\end{tabular}
\end{table}

Median: 15 minutes per drawing (range across participants: 10--180)

\FloatBarrier

\subsubsection*{Time Estimates for 100
Drawings}\label{time-estimates-for-100-drawings}
\addcontentsline{toc}{subsubsection}{Time Estimates for 100 Drawings}

\begin{table}[H]
\centering
\caption{Estimated time for processing 100 drawings: PyPottery vs.\ traditional workflow.}
\begin{tabular}{c c c c}
\toprule
\textbf{Participant} & \textbf{PyPottery (hours)} & \textbf{Traditional (hours)} & \textbf{Speedup Factor} \\
\midrule
0 & 1   & 120 & \textbf{120$\times$} \\
1 & 2   & 80  & \textbf{40$\times$} \\
2 & 1.5 & 150 & \textbf{100$\times$} \\
3 & 1.5 & 48  & \textbf{32$\times$} \\
4 & 4   & 70  & \textbf{17.5$\times$} \\
\bottomrule
\end{tabular}
\end{table}

Mean estimated speedup: \(\sim\)62\(\times\) faster than traditional
workflow.

\FloatBarrier

\subsubsection*{Greatest Time Savings Phase
(Open-Ended)}\label{greatest-time-savings-phase-open-ended}
\addcontentsline{toc}{subsubsection}{Greatest Time Savings Phase
(Open-Ended)}

\begin{table}[H]
\centering
\caption{Phase identified as providing the greatest time savings.}
\begin{tabular}{lc}
\toprule
\textbf{Phase} & \textbf{Votes} \\
\midrule
Inking/Vectorization & 4 \\
Scanning/OCR & 1 \\
Layout & 1 \\
\bottomrule
\end{tabular}
\end{table}

\emph{(Note: Participant 4 voted for both Scanning/OCR and Layout)}

\FloatBarrier

\subsubsection*{Phase Needing Most
Improvement}\label{phase-needing-most-improvement}
\addcontentsline{toc}{subsubsection}{Phase Needing Most Improvement}

\begin{table}[H]
\centering
\caption{Phase identified as needing the most improvement.}
\begin{tabular}{lc}
\toprule
\textbf{Phase} & \textbf{Votes} \\
\midrule
Inking/Vectorization & 4 \\
Scanning/OCR & 1 \\
\bottomrule
\end{tabular}
\end{table}

\FloatBarrier

\subsection{Task Completion \&
Efficiency}\label{task-completion-efficiency}

\FloatBarrier

\subsubsection*{Task Completion Status}\label{task-completion-status}
\addcontentsline{toc}{subsubsection}{Task Completion Status}

\begin{table}[H]
\centering
\caption{Task completion status across the six standardized tasks.}
\small
\begin{tabular}{l c c c c}
\toprule
\textbf{Task} & \textbf{Complete} & \textbf{Partial} & \textbf{Failed} & \textbf{Success Rate} \\
\midrule
T1 (Sheet Processing) & 5 & 0 & 0 & \textbf{100\%} \\
T2 (OCR Extraction)   & 4 & 1 & 0 & \textbf{80\%} \\
T3 (Parsing)          & 5 & 0 & 0 & \textbf{100\%} \\
T4 (Inking)           & 5 & 0 & 0 & \textbf{100\%} \\
T5 (Vectorization)    & 1 & 4 & 0 & \textbf{20\%} \\
T6 (Layout)           & 5 & 0 & 0 & \textbf{100\%} \\
\bottomrule
\end{tabular}
\end{table}

Overall Task Completion Rate: 83.3\% (25/30 tasks fully completed)

\FloatBarrier

\subsubsection*{Critical Observations}\label{critical-observations}
\addcontentsline{toc}{subsubsection}{Critical Observations}

\begin{itemize}
\item
  \textbf{T5 (Vectorization)} had the lowest completion rate with 4/5
  participants achieving only partial completion

  \begin{itemize}
  \item
    Primary issue: GUI difficulties with semantic-aware vectorization
  \item
    Multiple help requests recorded
  \end{itemize}
\item
  \textbf{T2 (OCR)} had 1 critical incident where OCR stopped working
  (resolved by restart)
\end{itemize}

\FloatBarrier

\subsubsection*{Help Requests by Task}\label{help-requests-by-task}
\addcontentsline{toc}{subsubsection}{Help Requests by Task}

\begin{table}[H]
\centering
\caption{Help requests recorded per task.}
\begin{tabular}{lc}
\toprule
\textbf{Task} & \textbf{Total Help Requests} \\
\midrule
T1 & 1 \\
T2 & 0 \\
T3 & 0 \\
T4 & 0 \\
T5 & \textbf{6} \\
T6 & 0 \\
\bottomrule
\end{tabular}
\end{table}

\FloatBarrier

\subsection{Qualitative Findings (Interview
Themes)}\label{qualitative-findings-interview-themes}

\FloatBarrier

\subsubsection*{Most Positive Feedback}\label{most-positive-feedback}
\addcontentsline{toc}{subsubsection}{Most Positive Feedback}

\begin{enumerate}
\def\labelenumi{\arabic{enumi}.}
\item
  GUI/UX Quality

  \begin{itemize}
  \item
    ``UX/UI well structured - Cool!''
  \item
    ``Great UI''
  \item
    ``Software aesthetics and usability''
  \end{itemize}
\item
  Archaeological Workflow Alignment

  \begin{itemize}
  \item
    ``Each step is well defined for archaeological purpose''
  \item
    ``The archaeological workflow is clear and well defined within the
    computational process''
  \item
    ``Built-in software for archaeology''
  \end{itemize}
\item
  Time Savings

  \begin{itemize}
  \item
    ``Extremely useful, great time saver''
  \item
    ``Unimaginable potential time savings''
  \item
    ``Automation of repetitive processes''
  \end{itemize}
\item
  Independence from External Software

  \begin{itemize}
  \item
    ``No dependence on other external software''
  \item
    ``{[}Could{]} completely replace my traditional workflow''
  \end{itemize}
\end{enumerate}

\FloatBarrier

\subsubsection*{Most Common Pain Point}\label{most-common-pain-point}
\addcontentsline{toc}{subsubsection}{Most Common Pain Point}

Installation Process --- cited by 3/5 participants as the most
confusing/difficult aspect.

\FloatBarrier

\subsubsection*{Feature Requests}\label{feature-requests}
\addcontentsline{toc}{subsubsection}{Feature Requests}

\begin{enumerate}
\def\labelenumi{\arabic{enumi}.}
\item
  Prospectus Vectorization (mentioned by 3 participants)
\item
  Database Integration in PyPotteryScan
\item
  Diameter Positioning with Measurement input
\item
  Improved Dotting/Stippling Control
\item
  Scale Calibration Method in PyPotteryLayout
\item
  AI Assistant for software questions
\item
  Stylistic Palette for graphic rendering
\end{enumerate}

\FloatBarrier

\subsubsection*{Net Promoter Score}\label{net-promoter-score}
\addcontentsline{toc}{subsubsection}{Net Promoter Score}

\begin{table}[H]
\centering
\caption{Net Promoter Score responses.}
\begin{tabular}{cl}
\toprule
\textbf{Score} & \textbf{Reason} \\
\midrule
10 & ``Speed'' \\
10 & ``Time saving to do other stuff'' \\
9  & ``Great job'' \\
9  & ``A lot of time saved'' \\
9  & ``Unimaginable potential time savings'' \\
\bottomrule
\end{tabular}
\end{table}

Average NPS: 9.4/10.

\section{Early Adoption: Applicability Across Additional
Contexts}\label{sec-appB}

This appendix documents seven contexts in which PyPottery is currently
being applied. These are the cases known to the author, the researchers
involved having made contact directly; as the software requires no
registration and operates entirely locally, the actual extent of use
cannot be determined.

Beyond the primary case study presented in Section 3.3 and evaluated in
Section 4, PyPottery is currently being applied by independent
researchers to archaeological material from a range of chronological,
geographical, and linguistic contexts. This appendix documents these
instances of early adoption as evidence of the software's applicability
beyond the Terramara di Montale case study. These cases are presented as
qualitative evidence of applicability and are not intended as a
quantitative validation of the results reported in Section 4; systematic
cross-site evaluation, involving formal usability and performance
metrics analogous to those applied to the Montale dataset, is identified
as a priority for future community-driven work (see Section 5.4).

These researchers were not recruited for this study. Each began using
PyPottery through the openly available repository, before any
peer-reviewed publication of the suite, having encountered it through
disciplinary circulation or personal communication. The author's contact
with them was limited to requesting consent for their inclusion here and
collecting the descriptive information reported in the following table.
It remains an early and limited stage of dissemination, and we present
it as such: seven contexts do not constitute a validation sample, and
Section 4 remains the manuscript's only formal evaluation.

The table summarizes the seven contexts, spanning Late Mesolithic to
Early Medieval chronologies, Italian, German, and mixed English/Italian
documentation practices, and coarse to fine ceramic fabrics. Researchers
are credited with their consent; all data shown (context, chronology,
material, modules used) were provided directly by the researchers and
reflect their own characterization of their ongoing work.

\begin{table}[H]
\centering
\caption{Early adoption contexts.}
\scriptsize
\resizebox{\textwidth}{!}{%
\begin{tabular}{L{2.4cm} L{2.6cm} L{2cm} L{1.9cm} L{1.4cm} L{1.6cm} L{1.6cm}}
\toprule
\textbf{Context} & \textbf{Researcher / Institution} & \textbf{Chronology} & \textbf{Material} & \textbf{Doc. language} & \textbf{Modules used} & \textbf{Status} \\
\midrule
Pyrgi, ``public-ceremonial'' quarter (Santa Severa, Santa Marinella, Rome) & S. Servoli, Dipartimento di Scienze dell'Antichit\`a, Sapienza University of Rome & Late 7th -- late 5th c. BCE & ``rosso-bruno'' impasto ware & Italian & Scan, Ink, Trace, Layout & Ongoing preparation of doctoral monograph \\
\addlinespace
Biriai (Oliena, Nuoro, Sardegna), Monte Claro culture settlement & F.I. Debandi, PhD, Post-doc fellow, Dept.\ of History and Cultures (DISCI), University of Bologna & Copper Age (2700--2500 BCE) & Impasto ceramic & Italian & Scan, Ink & Material study \\
\addlinespace
H\"ude I, distr.\ Diepholz & A. Philippi, LWL-Arch\"aologie f\"ur Westfalen / Au\ss enstelle Bielefeld & Late Mesolithic -- Early Neolithic (Bischheim, Swifterbant, Michelsberg, Trichterbecher; c.\ 1{,}500-year span) & Impasto ceramic & German & Ink, Layout & Material study \\
\addlinespace
Torre dell'Alto, Nard\`o (LE) & L. Schiavone, Universit\`a di Bologna & Mid--Late Bronze Age & Impasto ware & Italian & Ink, Layout & Ongoing paper publication \\
\addlinespace
Arslantepe, ``Edificio XLVI'' & V. Parisi, PhD Program in Heritage Science, Dept.\ of Sciences of Antiquity, Sapienza University of Rome & Iron Age II/Middle (1000--850 BCE) & --- & English/Italian & Ink, Trace, Layout & Material study \\
\addlinespace
Shamiram-Aruch (Aragatsotn, Republic of Armenia) & E. Fausti, Dipartimento di Scienze dell'Antichit\`a, Sapienza University of Rome & Middle Iron Age -- Early Medieval (6th c.\ BCE--7th c.\ CE) & Impasto, depurated ware & Italian & Trace, Ink, Layout & Ongoing preparation of several scientific papers \\
\addlinespace
San Vincenzo, Stromboli (ME) & A. Di Renzoni, CNR ISPC & Middle Bronze Age & Impasto ware & Italian & Scan, Ink, Layout & Material study \\
\bottomrule
\end{tabular}%
}
\end{table}

\begin{figure}

\centering{

\includegraphics[width=0.85\linewidth,height=\textheight,keepaspectratio]{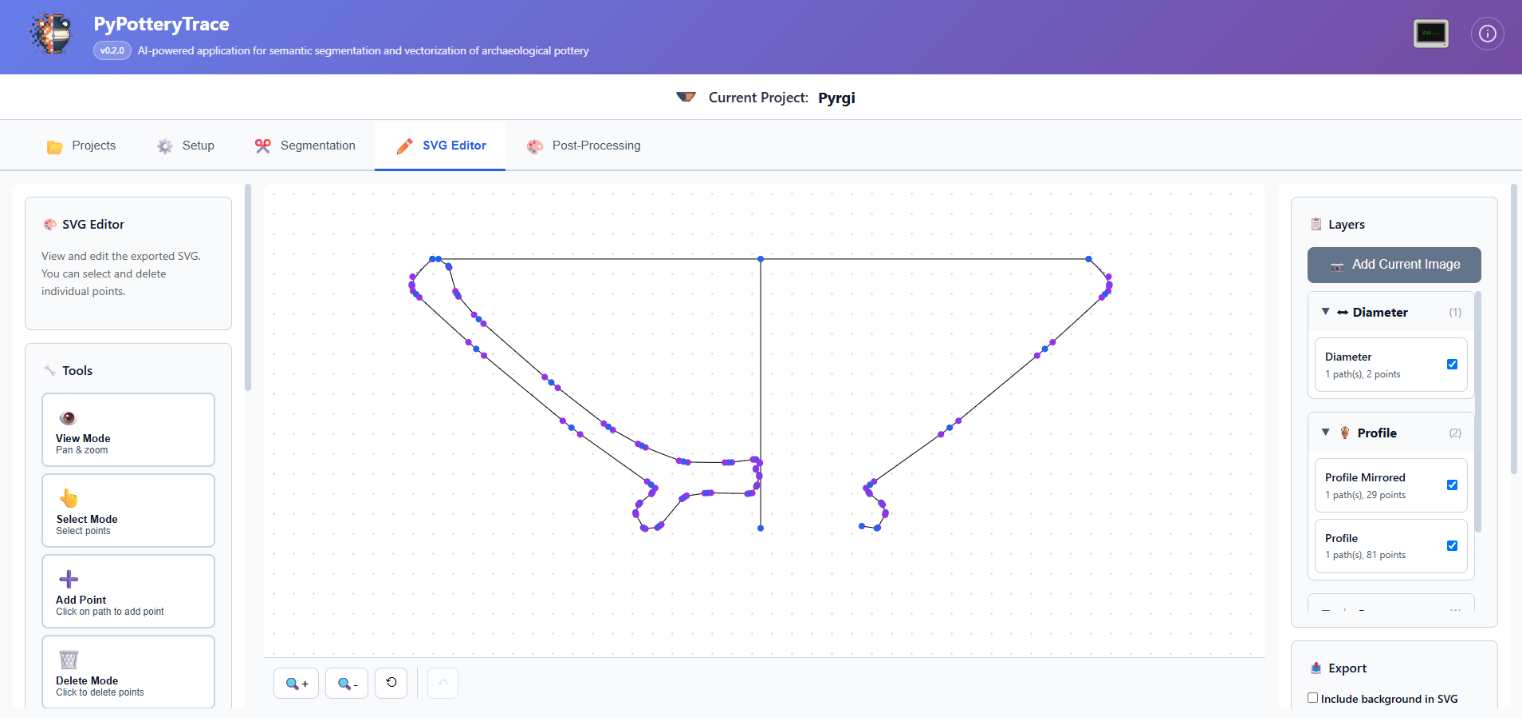}

}

\caption{\label{fig-SB1}Pyrgi (Santa Severa, Rome). PyPotteryTrace GUI
applied to ``rosso-bruno'' impasto ware.}

\end{figure}%

\begin{figure}

\centering{

\includegraphics[width=0.8\linewidth,height=\textheight,keepaspectratio]{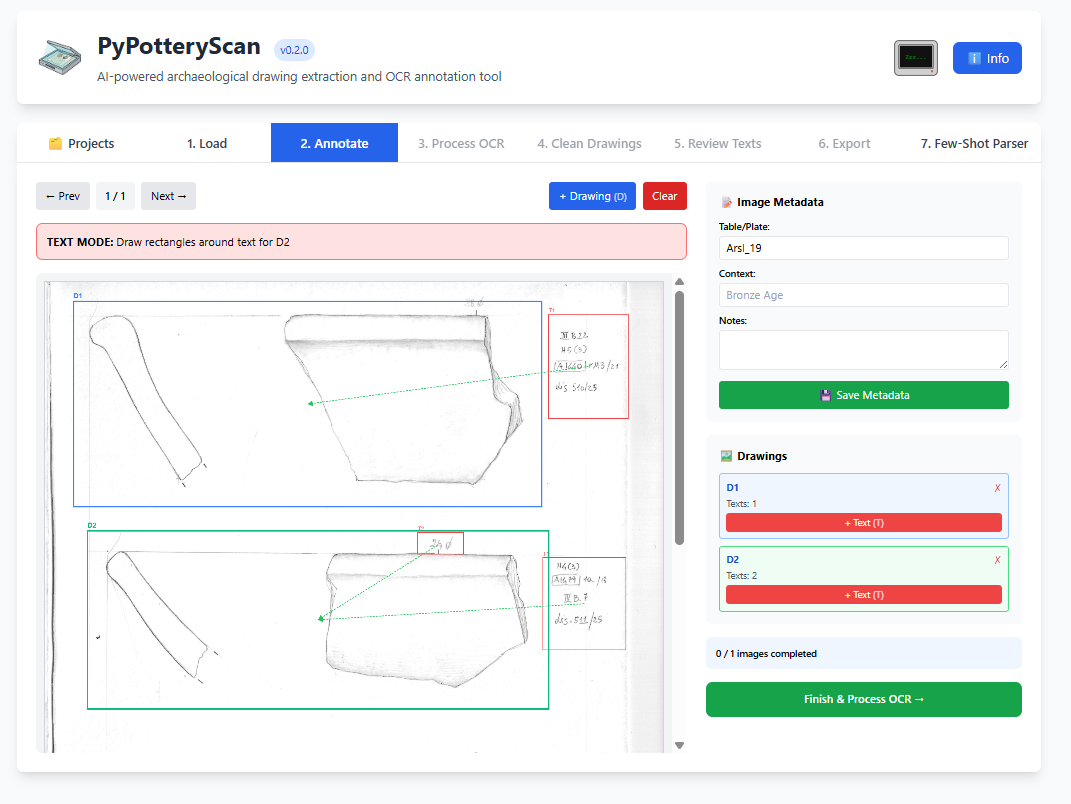}

}

\caption{\label{fig-SB2}Arslantepe, ``Edificio XLVI'' --- PyPotteryScan
GUI.}

\end{figure}%

\begin{figure}

\centering{

\includegraphics[width=0.85\linewidth,height=\textheight,keepaspectratio]{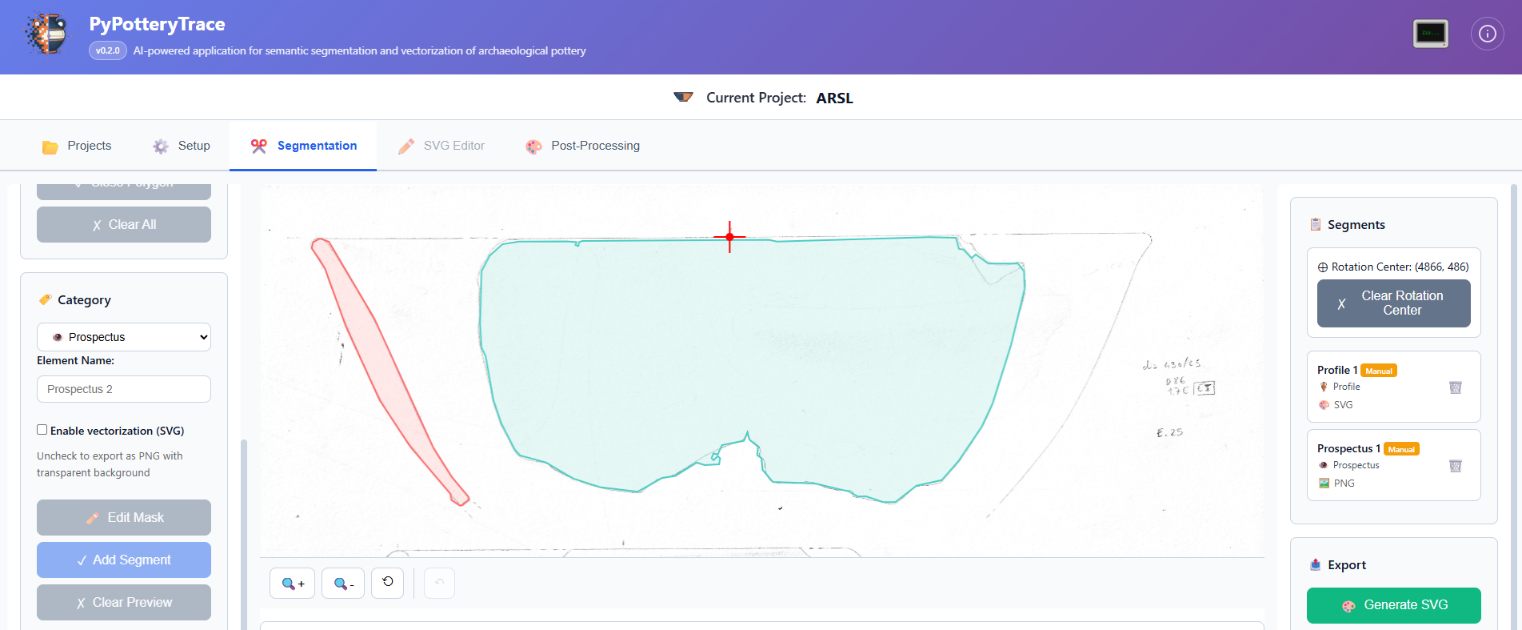}

}

\caption{\label{fig-SB3}Arslantepe, ``Edificio XLVI'' ---
PyPotteryTrace.}

\end{figure}%

\begin{figure}

\centering{

\includegraphics[width=0.8\linewidth,height=\textheight,keepaspectratio]{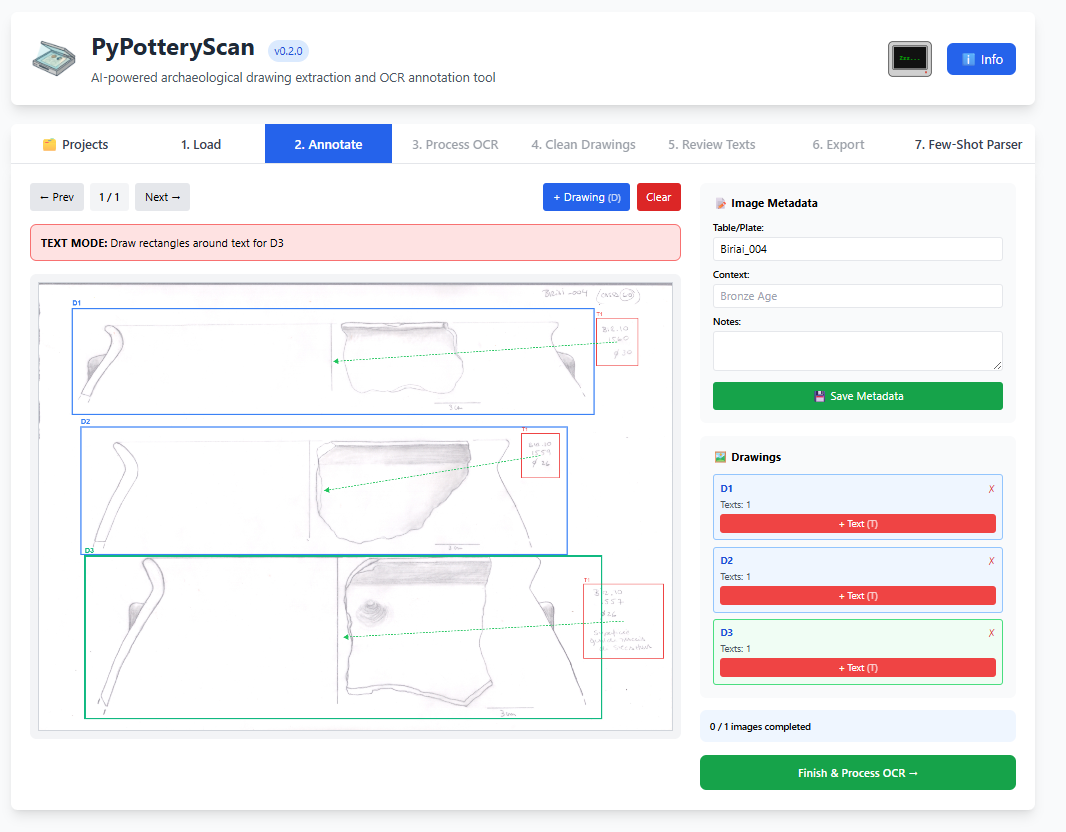}

}

\caption{\label{fig-SB4}Biriai (Oliena, Sardegna) --- PyPotteryScan
GUI.}

\end{figure}%

\begin{figure}

\centering{

\includegraphics[width=0.8\linewidth,height=\textheight,keepaspectratio]{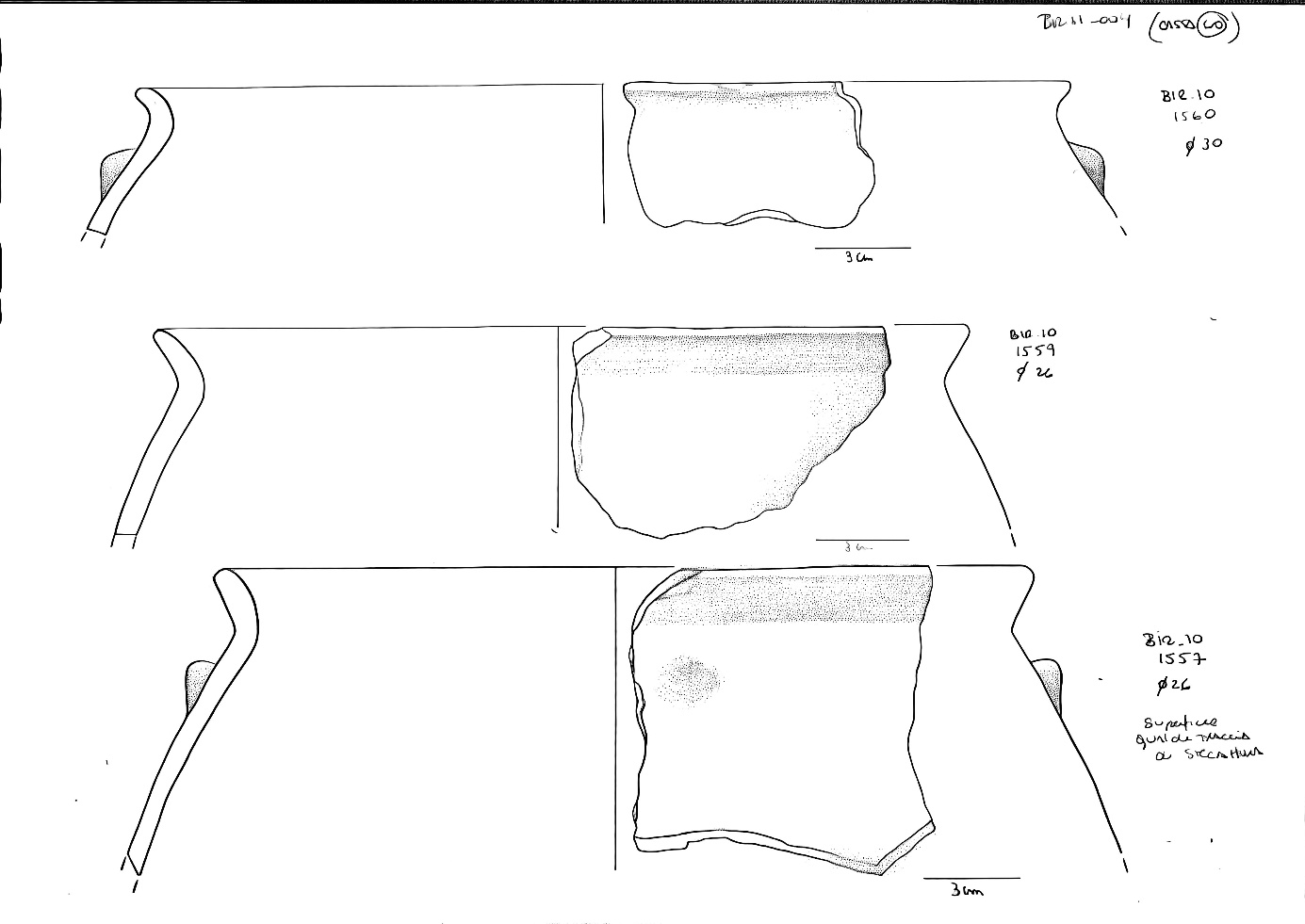}

}

\caption{\label{fig-SB5}Biriai (Oliena, Sardegna) --- PyPotteryInk
output inked plate.}

\end{figure}%

\begin{figure}

\centering{

\includegraphics[width=0.6\linewidth,height=\textheight,keepaspectratio]{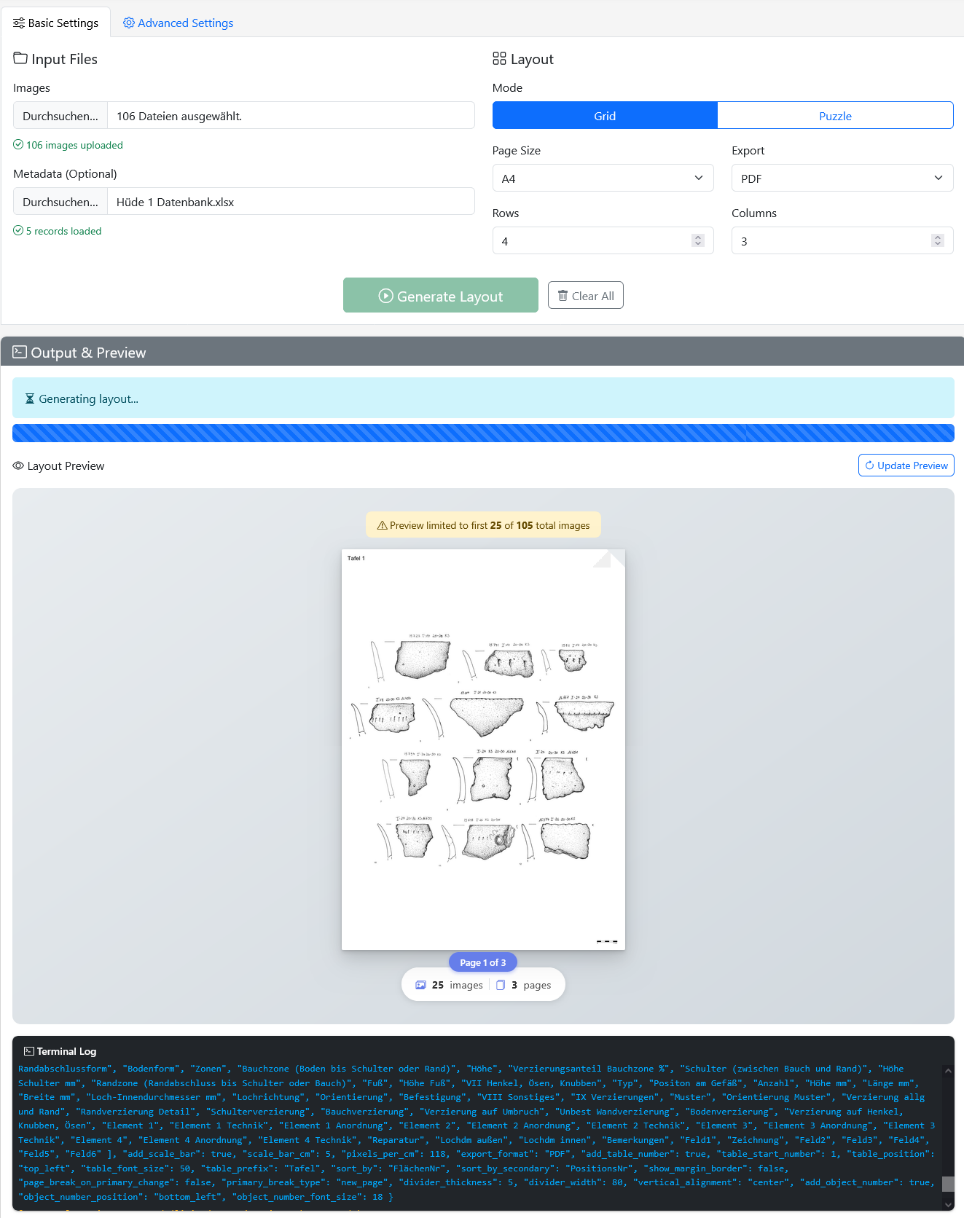}

}

\caption{\label{fig-SB6}Hüde I (Diepholz) --- PyPotteryLayout GUI.}

\end{figure}%

\begin{figure}

\centering{

\includegraphics[width=0.55\linewidth,height=\textheight,keepaspectratio]{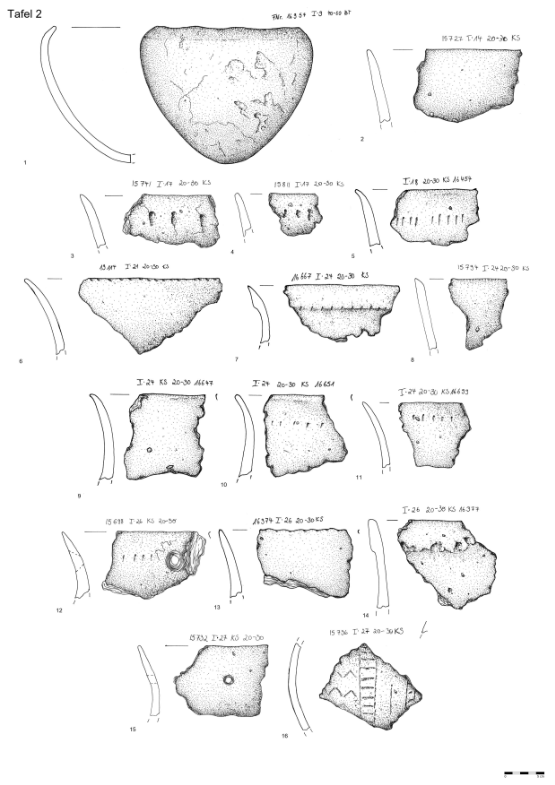}

}

\caption{\label{fig-SB7}Hüde I (Diepholz) --- PyPotteryLayout output.}

\end{figure}%

\begin{figure}

\centering{

\includegraphics[width=0.65\linewidth,height=\textheight,keepaspectratio]{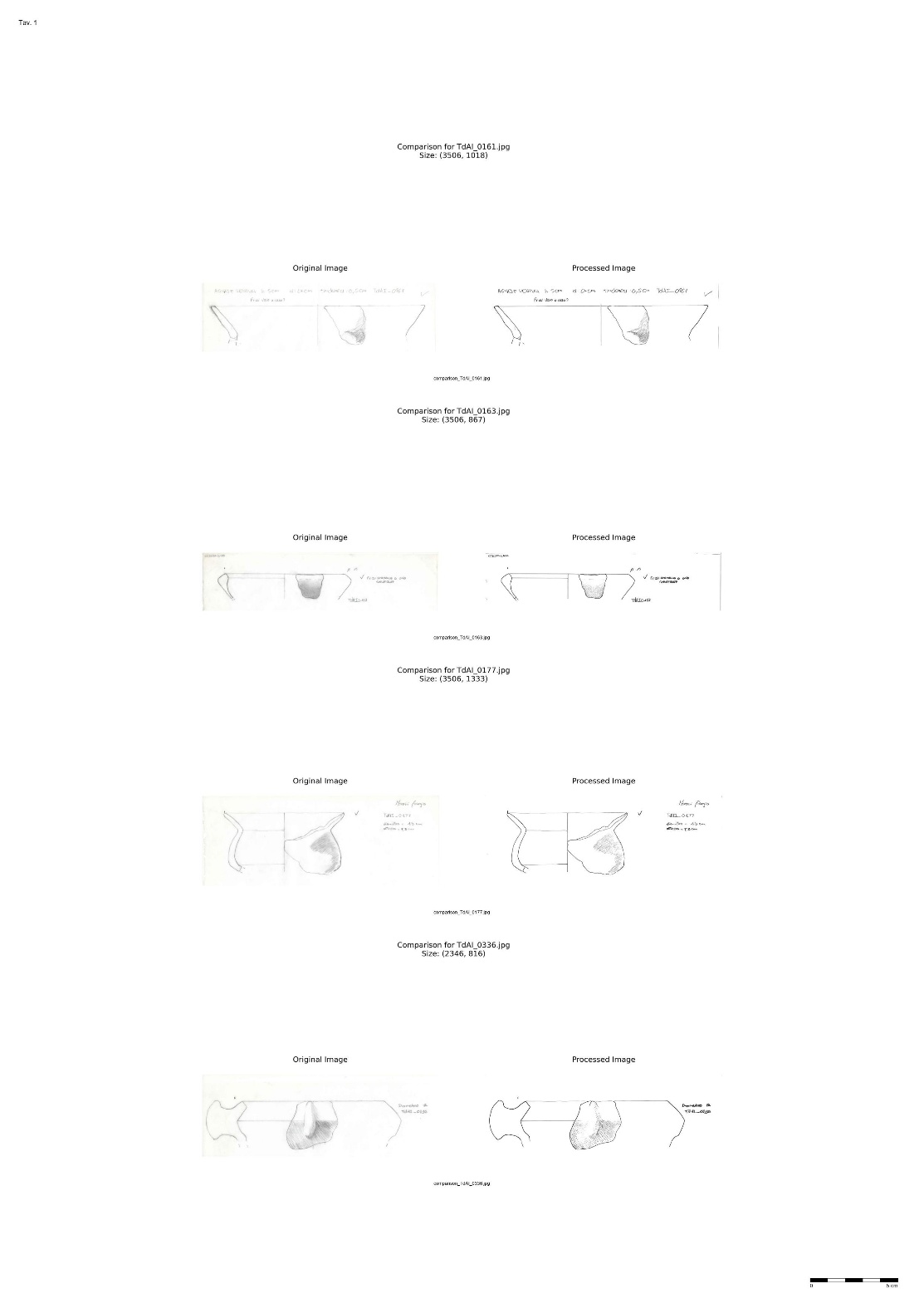}

}

\caption{\label{fig-SB8}Torre dell'Alto, Nardò --- comparison plate
between original and PyPotteryInk-processed drawings.}

\end{figure}%

\begin{figure}

\centering{

\includegraphics[width=0.85\linewidth,height=\textheight,keepaspectratio]{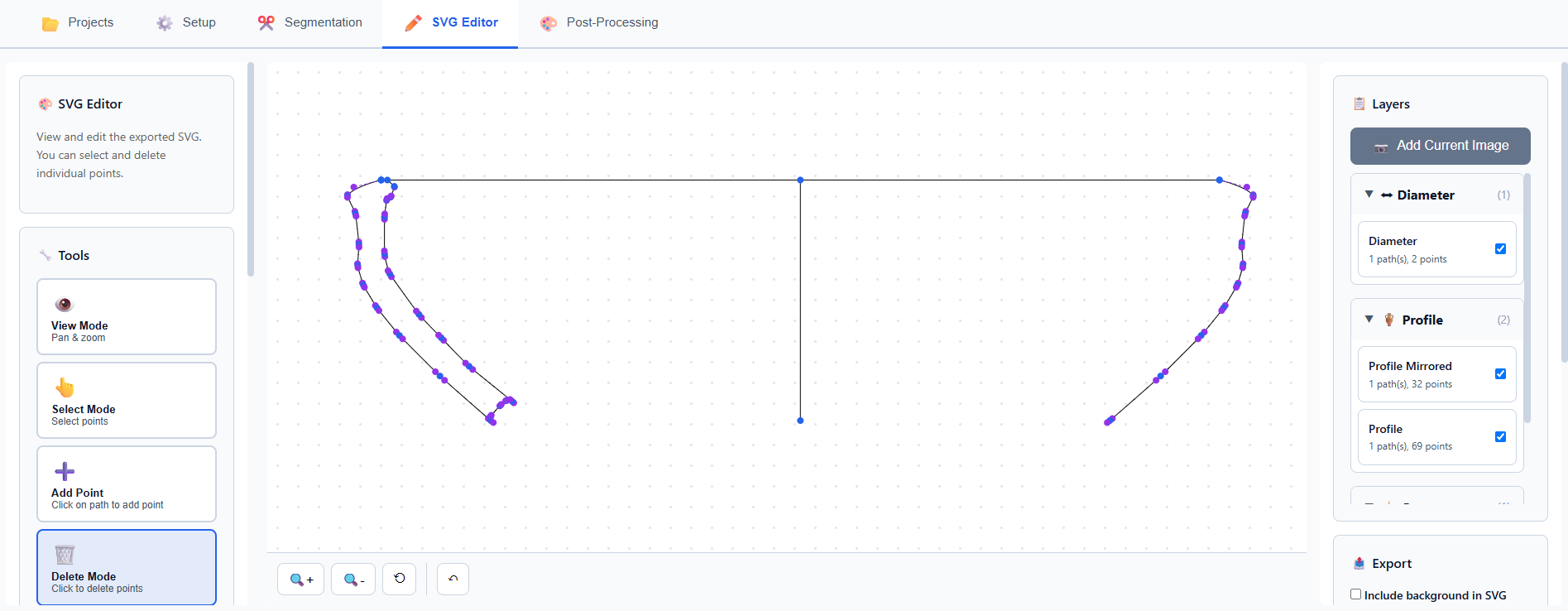}

}

\caption{\label{fig-SB9}Shamiram-Aruch (Armenia) --- PyPotteryTrace
GUI.}

\end{figure}%

\begin{figure}

\centering{

\includegraphics[width=0.85\linewidth,height=\textheight,keepaspectratio]{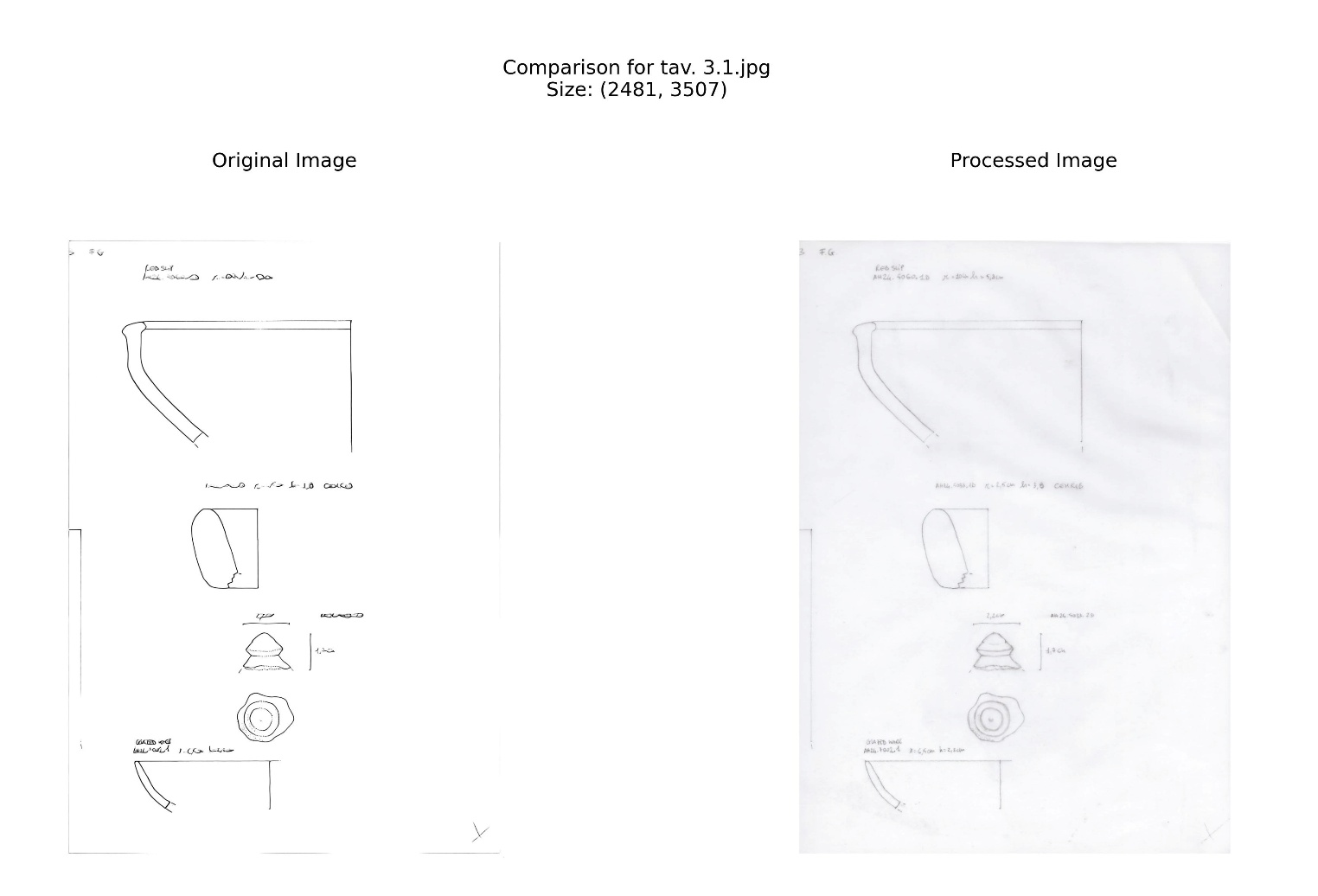}

}

\caption{\label{fig-SB10}Shamiram-Aruch (Armenia) --- comparison plates
between original and Ink-processed drawings.}

\end{figure}%

\begin{figure}

\centering{

\includegraphics[width=0.85\linewidth,height=\textheight,keepaspectratio]{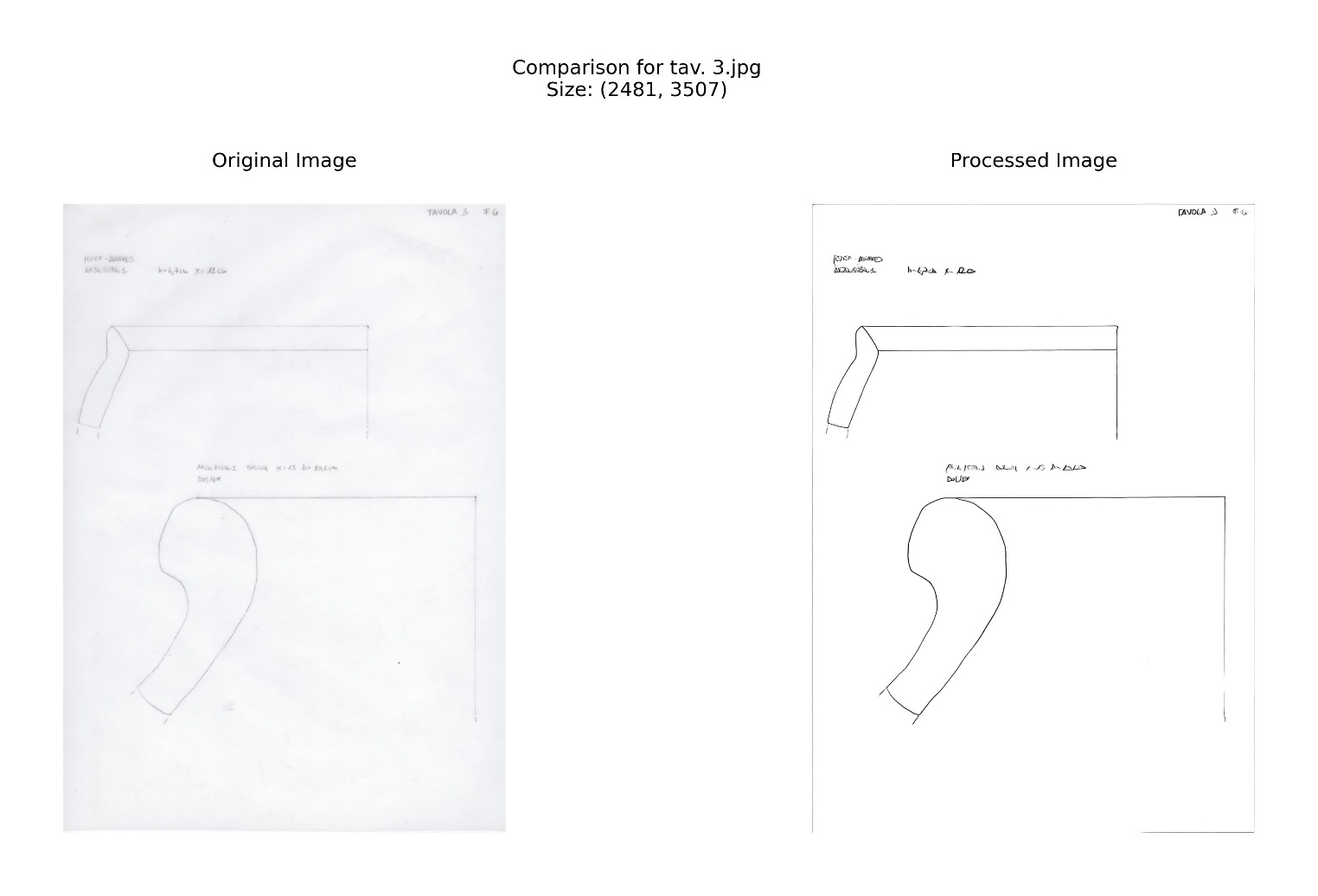}

}

\caption{\label{fig-SB11}Shamiram-Aruch (Armenia) --- comparison plates
between original and Ink-processed drawings.}

\end{figure}%

\end{document}